\documentclass{article}
\usepackage{iclr2027_conference,times}
\usepackage[hyphens]{url}
\usepackage{hyperref}
\usepackage{graphicx}
\usepackage{subcaption}
\usepackage{booktabs}
\usepackage{multirow}
\usepackage{algorithm}
\usepackage{algorithmic}
\usepackage{amsmath}
\usepackage{amssymb}
\usepackage{mathtools}
\usepackage{amsthm}
\usepackage{placeins}
\usepackage{tcolorbox}
\tcbuselibrary{breakable,skins}
\usepackage{listings}

\theoremstyle{plain}

\theoremstyle{definition}

\theoremstyle{remark}

\newif\ifarxivversion
\arxivversiontrue 

\title{Bridging Exploration and Persona: Information-Directed Dueling for Test-Time Agent Personalization}
\ifarxivversion
  \hypersetup{
    pdftitle={Bridging Exploration and Persona: Information-Directed Dueling for Test-Time Agent Personalization},
    pdfauthor={Jie Mao, Pingchen Lu, Xiangyi Wang, Hengquan Huang, Shuang Qiu, Min Zhang, Zhongxiang Dai}}
  \author{
    \normalfont\textbf{Jie Mao}\textsuperscript{1}, Pingchen Lu\textsuperscript{1},
    Xiangyi Wang\textsuperscript{1}, Hengquan Huang\textsuperscript{2},\\
    Shuang Qiu\textsuperscript{3}, Min Zhang\textsuperscript{4,*},
    Zhongxiang Dai\textsuperscript{1,*}\\
    {\normalfont\small \textsuperscript{1}The Chinese University of Hong Kong, Shenzhen}\\
    {\normalfont\small \textsuperscript{2}University of Copenhagen}\\
    {\normalfont\small \textsuperscript{3}City University of Hong Kong}\\
    {\normalfont\small \textsuperscript{4}East China Normal University}\\
    {\normalfont\small \textsuperscript{*}Corresponding authors}\\
    {\normalfont\small Code: \url{https://github.com/s1412/ID-TAP}}}
  \iclrfinalcopy
\else
  \hypersetup{
    pdftitle={Bridging Exploration and Persona: Information-Directed Dueling for Test-Time Agent Personalization},
    pdfauthor={Anonymous authors}}
  \author{Anonymous authors\\Paper under double-blind review}
\fi

\begin{document}
\maketitle
\ifarxivversion
  \lhead{Preprint}
\fi

\begin{abstract}
Effective test-time personalization requires LLM agents to adapt to query-dependent user preferences from lightweight interaction. We study persona selection with binary comparisons between candidate responses as the online signal. We present ID-TAP, a scalable, IDS-inspired neural dueling-bandit framework for large discrete persona spaces. ID-TAP combines a query-conditioned neural utility model with a lightweight Bradley--Terry explorer, and transfers comparison evidence across related queries through similarity-weighted warm starts. The neural model shares statistical strength across queries while preserving query-specific scores, and efficient BT updates score hypothetical challengers. Across seven user-conditioned generation benchmarks under reference- and model-based preference simulators, ID-TAP achieves the strongest aggregate performance among the evaluated baselines within a fixed ten-round comparison budget. These results support information-directed challenger selection as an effective design for test-time persona adaptation.
\end{abstract}

\section{Introduction}
Large language model (LLM)-based agents support tasks ranging from creative writing to coding assistance~\citep{huang2023agentcoder,li2024agent}. Their behavior, including response style and persona, is strongly influenced by the instructions supplied at inference time. Prompt design and automatic instruction optimization improve task-level behavior~\citep{wei2022chain,zhou2022large}; personalization extends this objective toward responses suited to an individual user~\citep{salemi2024lamp}.

A natural control interface is the agent's \textit{persona}: a prompt-level description of the behavior and style the agent should adopt. Persona selection differs from task-level prompt optimization in three ways.
First, personalization is \textbf{contextual}: the persona most useful for one request may be less suitable for another (e.g., a ``creative collaborator'' for brainstorming versus a ``rigorous critic'' for editing)~\citep{salemi2024lamp,zhang2025personaagent}.
Second, user preferences can be \textbf{implicit and noisy}: interaction histories provide partial evidence about relevant tendencies~\citep{richardson2023integrating}.
Third, candidate persona descriptions can be \textbf{semantically dense}. Closely related descriptions (e.g., ``stern tutor'' and ``strict mentor'') may induce meaningfully different responses despite high embedding similarity.

These properties motivate \textbf{test-time adaptation (TTA)} over a fixed persona pool. Prompt-level adaptation preserves a shared backbone while incorporating feedback about the current request. PersonaAgent~\citep{zhang2025personaagent} uses expressive linguistic feedback, while \textbf{preference feedback} reduces an interaction to choosing between two candidate responses, a supervision format also used in RLHF~\citep{ouyang2022training}.

Recent approaches such as POHF~\citep{lin2024prompt} and PDO~\citep{wu2025llm} apply bandit exploration to preference-driven prompt optimization. In the persona pools evaluated here, their UCB- and Thompson-sampling-based components revisit challengers more often and reach lower aggregate performance than our information-directed rule (Sec.~\ref{sec:experiment}).

Building on information-directed and dueling-bandit methods, we develop ID-TAP (\textbf{I}nformation-Directed \textbf{D}ueling for \textbf{T}est-Time \textbf{A}gent \textbf{P}ersonalization), a scalable challenger-selection framework that uses pairwise preferences as its online signal. Its technical focus is the joint design required by black-box personalization over large persona pools: query-conditioned utility estimation, lightweight hypothetical BT updates, and transfer across related queries.
Specifically, a query-conditioned neural utility model estimates utility gaps, while a lightweight statistical explorer evaluates hypothetical comparison outcomes through an IDS-inspired entropy-reduction proxy. This separation supports efficient challenger scoring with one neural update per observed comparison, and similarity-weighted warm starts reuse preference evidence across queries. Under a fixed comparison budget, the resulting method attains the strongest aggregate result among the evaluated baselines.

Our contributions are summarized as follows:

\paragraph{Scalable Challenger Acquisition.} We operationalize an IDS-inspired acquisition criterion for large discrete persona pools by combining neural utility gaps with a lightweight BT entropy-reduction proxy, enabling efficient scoring across candidate challengers.
\paragraph{Cross-Query Adaptation.} We introduce a similarity-weighted warm start that reuses comparison evidence across related requests while retaining query-conditioned utility prediction.
\paragraph{Empirical Evaluation.} Across seven benchmarks and two offline preference simulators, ID-TAP achieves the strongest aggregate result among the compared methods under the same ten-round budget. On the aggregate generation-task trajectory, it overtakes the strongest compared baseline after approximately \textbf{4 simulated-feedback rounds}, suggesting favorable feedback-round efficiency in this setting.

\begin{figure}[t]
\centering
\includegraphics[width=\linewidth]{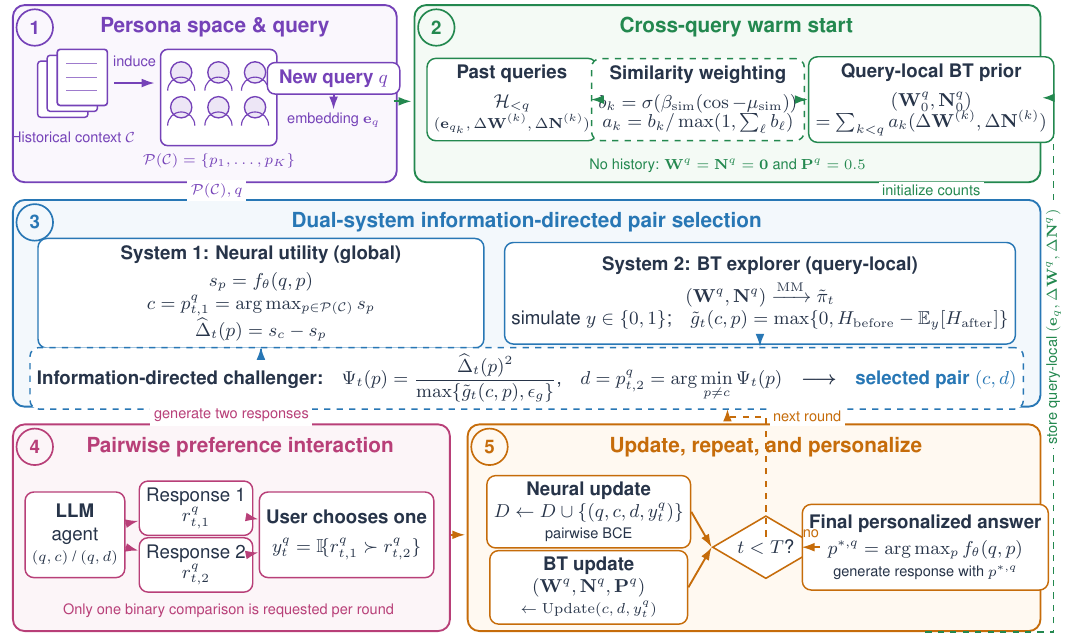}
\caption{\textbf{Overview of ID-TAP.} Historical context induces a shared persona space and each new query inherits similarity-weighted BT statistics from prior queries (1--2). At every interaction round, the neural utility model selects the champion and estimates utility gaps, while the query-local BT explorer computes a plug-in entropy-reduction proxy; their acquisition ratio selects the challenger (3). The agent generates the resulting response pair, obtains one binary preference, and updates both systems (4--5). After $T$ rounds, the highest-scoring persona produces the personalized response, and the query-local BT increments are retained for later warm starts.}
\label{fig:framework}
\end{figure}

\section{Problem Formulation}
\label{sec:pf}

We formalize persona selection as a contextual, query-level problem with an offline construction phase and an online test-time adaptation phase.

\subsection{Persona and Contextual Space}
For a personalized generation task, a \textit{persona} $p$ is a prompt-level description that summarizes relevant traits, stylistic habits, and task preferences inferred from the user's historical interaction context $\mathcal{C}$.
The offline procedure constructs a finite persona space $\mathcal{P}(\mathcal{C})=\{p_1,\ldots,p_K\}$ from this history. The candidates capture different emphases that remain grounded in $\mathcal{C}$. Because requests from the same user can differ in intent and complexity, their utilities are query-dependent rather than governed by a single universally best candidate.

\subsection{Offline--Online Framework}
We use the following offline--online protocol:

\paragraph{Offline Phase (Persona Induction).} Historical context $\mathcal{C}$ is used to construct the shared candidate set $\mathcal{P}(\mathcal{C})$. The candidate strings are fixed during online adaptation and can be reused across the user's subsequent queries.

\paragraph{Online Phase (Query-Specific Adaptation).} At test time, the agent receives a query $q$. Let $u(q,p)$ denote the unknown expected utility of the response generated under persona $p$. The query-conditioned optimal candidate is
\begin{equation}
    p^{*,q} = \arg\max_{p \in \mathcal{P}(\mathcal{C})} u(q,p).
\end{equation}
The neural score $f_\theta(q,p)$ introduced in Section~\ref{sec:method} estimates this latent utility from observed comparisons.

\subsection{Test-Time Adaptation with Preference Feedback}

ID-TAP learns estimates of the latent utility online while keeping the finite candidate personas fixed. We assume that user preferences are locally stable over short interaction horizons. Test-time adaptation therefore operates within the pre-constructed persona space $\mathcal{P}(\mathcal{C})$, with the historical context $\mathcal{C}$ fixed throughout the session.

To identify a strong persona from lightweight interaction, we adopt a dueling-bandit interface in which the agent learns from pairwise preference feedback. At interaction step $t$, the algorithm selects a champion persona and a challenger persona, then generates the corresponding response pair $(r_t^1,r_t^2)$. The champion exploits the neural model's current estimate, whereas the challenger is chosen by the IDS-inspired acquisition rule in Section~\ref{sec:method}. The user returns a binary comparison $y_t=\mathbb{I}\{r_t^1\succ r_t^2\}$, which updates both utility estimates and pairwise statistics. Given an interaction budget $T$, the objective is to return a persona $\hat p^q_T$ with high query-conditioned utility from these binary comparisons.

\section{Methodology}
\label{sec:method}
\begin{algorithm}[t]
    \begin{algorithmic}[1]
    
    \STATE \textbf{Initialize} Neural network $f_\theta$, persona space $\mathcal{P}(\mathcal{C})$
        \FOR{$q=1,2,\ldots, N$}
            \STATE Initialize $(\mathbf{W}^{q},\mathbf{N}^{q},\mathbf{P}^{q})$ by Algorithm~\ref{algo:inheritance}; set $(\Delta\mathbf W^q,\Delta\mathbf N^q)\leftarrow(\mathbf0,\mathbf0)$
            \FOR{$t=1,2,\ldots, T$}
                
                \STATE $p_{t,1}^{q} \leftarrow \arg\max\limits_{p \in \mathcal{P}(\mathcal{C})} f_\theta(q,p)$
                \STATE $p_{t,2}^{q} \leftarrow 
                    \arg\min\limits_{p \in \mathcal{P}(\mathcal{C})\setminus\{p_{t,1}^q\}}
                    \frac{\hat{\Delta}_{t}(p)^2}
                    {\max\{\tilde g_{t}(p_{t,1}^{q},p),\epsilon_g\}}$
                \STATE Generate response pair $\{r_{t,1}^{q},r_{t,2}^{q}\}$ 
                \STATE Observe preference feedback $y_{t}^{q} \leftarrow  \mathbb{I}\{r_{t,1}^{q} \succ r_{t,2}^{q}\}$ 
                \STATE Train $f_\theta$ with $D_t^q=\bigcup_{k<q}D_T^{q_k}\cup\{(q,p_{s,1}^{q},p_{s,2}^{q},y_s^q)\}_{s\leq t}$
                \STATE Update $(\mathbf{W}^{q},\mathbf{N}^{q},\mathbf{P}^{q})$ and local increments $(\Delta\mathbf W^q,\Delta\mathbf N^q)$
            \ENDFOR
            \STATE Store $(q,\mathbf e_q,\Delta\mathbf W^q,\Delta\mathbf N^q)$; output $\hat p_T^q \leftarrow \arg\max_{p} f_\theta(q,p)$
        \ENDFOR
    \caption{ID-TAP}   
    \label{algo}    
    \end{algorithmic}
\end{algorithm}

ID-TAP uses an \textbf{IDS-inspired acquisition} strategy for neural dueling bandits in a personalized, test-time interaction setting.
Before test time, we assume that the historical context set $\mathcal{C}$ and the corresponding persona space $\mathcal{P}(\mathcal{C})$ have been constructed offline. Inspired by \textbf{Information-Directed Sampling (IDS)}, our method balances exploitation and exploration by selecting comparison pairs that are both performance-relevant and uncertainty-reducing.
Specifically, the challenger is chosen by relating its estimated utility gap to a plug-in entropy-reduction proxy computed from the BT explorer.

\subsection{Algorithm Overview}
At test time, a user sequentially queries an AI agent with personalized requests. For each query $q$, the agent interacts with the user for a fixed budget of $T$ rounds, collecting pairwise feedback to adapt persona selection to the query-conditioned preference. The neural parameters persist across a user's queries; the BT state is initialized by Algorithm~\ref{algo:inheritance} and then updated locally. For the first query, the BT counts are zero and its pairwise probabilities are neutral. We use a fixed candidate-index rule to resolve exact ties in champion or challenger scores.

We model the utility function with a neural network. At each round, the agent first selects an exploitation arm
\begin{equation}
p_{t,1}^q = \arg\max_{p \in \mathcal{P}(\mathcal{C})} f_\theta(q, p),
\end{equation}
corresponding to the currently estimated best persona.

To efficiently explore alternative personas and reduce uncertainty, the second arm $p_{t,2}^q$ is selected using an information-directed criterion.
Concretely, we choose the challenger to minimize the acquisition ratio:
\begin{equation}
p_{t,2}^q = \arg\min\limits_{p \in \mathcal{P}(\mathcal{C})\setminus\{p_{t,1}^q\}}\frac{\hat{\Delta}_t(p)^2}{\max\{\tilde g_t(p_{t,1}^q,p),\epsilon_g\}},
\end{equation}
where $\hat{\Delta}_t(p) = f_\theta(q,p_{t,1}^q)-f_\theta(q,p)$ is the estimated utility gap, $\tilde g_t$ is the BT-based entropy-reduction proxy defined below, and $\epsilon_g>0$ prevents division by zero. The ratio favors challengers with a small predicted utility loss and a large estimated reduction in ranking uncertainty.

The user then provides preference feedback in the form of a binary comparison outcome. All collected pairwise interactions across queries and rounds are aggregated and used to update the neural utility model. After round $T$, the highest-scoring persona $\hat p_T^q$ conditions the final generated response.

\subsection{IDS-Inspired Challenger Selection}
Building on information-directed methods for bandit and preference feedback~\citep{russo2018learning,kirschner2020information,kirschner2021bias,pasztor2024bandits}, \textbf{ID-TAP} adapts the information-ratio principle to persona-level pair selection. We fix $p_{t,1}^q$ as the current greedy champion and select a challenger $p_{t,2}^q$ by minimizing the acquisition ratio.

\paragraph{Plug-in Entropy-Reduction Proxy.}
The BT solver produces strengths $\hat\theta_t^q$, which we map to the normalized plug-in score distribution $\tilde\pi_t^q(k)=\operatorname{softmax}(\hat\theta_t^q/\tau)_k$. For a champion--challenger pair $(p_{t,1}^q,p)$, we define
\begin{equation}
\label{eq:entropy_reduction_proxy}
\tilde g_t(p_{t,1}^q,p)=
\max\!\left\{0, H(\tilde\pi_t^q)
-\sum_{y\in\{0,1\}}\hat P_t(y\mid p_{t,1}^q,p)
H(\tilde\pi_{t}^{q,(y)})\right\},
\end{equation}
where $H(\tilde\pi)=-\sum_k\tilde\pi(k)\log\tilde\pi(k)$, $\hat P_t$ is the BT predictive probability, and $\tilde\pi_t^{q,(y)}$ is obtained by adding hypothetical outcome $y$ and re-estimating the BT strengths. This IDS-inspired plug-in score ranks pairs by their predicted reduction in the dispersion of normalized BT strengths.
Reproducible implementation details are provided in Appendix~\ref{app:bt_mi_impl}.

\subsection{Dual-System Architecture for Real-Time Efficiency}
Equation~(\ref{eq:entropy_reduction_proxy}) evaluates two hypothetical outcomes for each candidate, yielding $2(K-1)$ BT re-estimations per interaction round. The lightweight BT solver makes this candidate-wide scoring practical at large $K$, while the neural utility model performs one global update after each observed comparison. We therefore decouple global utility learning from local uncertainty estimation:
\begin{itemize}
\item \textbf{System 1: Neural Utility Model ($f_\theta$)}
This system serves as the shared utility estimator. It maps queries and persona embeddings to utility scores $s_p$ and provides the gap estimates $\hat{\Delta}_t(p)$ from the interaction history $D$ accumulated across queries. It is updated once per interaction round after observing feedback.
\item \textbf{System 2: Statistical Explorer (BT-MLE)} \\
This system acts as the query-local explorer. It maintains a lightweight Bradley--Terry (BT) model using comparison counts $\{w_{ij}, n_{ij}\}$. Its minorization--maximization (MM) solver can simulate the $2(K-1)$ hypothetical outcomes and recompute the normalized strength distribution efficiently.
\end{itemize}

\paragraph{Pairwise Probabilities.}
The BT explorer applies symmetric Beta smoothing to observed pairs and assigns neutral probability $P_{ij}=0.5$ to unobserved pairs. Appendix~\ref{app:bt_mi_impl} gives the count updates, probability estimator, and MM solver.

\paragraph{Design Motivation.}
The dual-system design combines complementary estimation properties. System~1 (the neural utility model $f_\theta$) is trained with pairwise BCE, corresponding to the negative BTL log-likelihood used by System~2 (BT-MLE). System~1 selects the champion and estimates utility gaps through shared representations, while System~2 supplies the challenger-side entropy-reduction proxy from pairwise outcomes. Consequently, $\Psi_t(p)=\hat{\Delta}_t(p)^2/\max\{\tilde g_t(p_{t,1}^q,p),\epsilon_g\}$ selects challengers that are both promising and informative. Existing neural-dueling, Bradley--Terry estimation, and classical M-estimation analyses motivate these components~\citep{verma2025neuralduelingbanditspreferencebased,hunter2004mm,van2000asymptotic,shah2015estimationpairwisecomparisonssharp}; our empirical study evaluates their coupled use with cross-query warm starts.

\paragraph{Cross-Query Knowledge Transfer.}
For a new query $q$, ID-TAP warm-starts the BT system from the query-local count increments $(\Delta\mathbf W^{(k)},\Delta\mathbf N^{(k)})$ collected on earlier queries. Let $b_k$ be the raw similarity score and normalize it as
\begin{equation}
\begin{aligned}
b_k&=\sigma\!\left(\beta_{\mathrm{sim}}(\mathrm{sim}(q,q_k)-\mu_{\mathrm{sim}})\right),\\
a_k&=\frac{b_k}{\max\{1,\sum_{\ell<q}b_\ell\}},\qquad \sum_{k<q}a_k\leq1,\\
(\mathbf W_0^q,\mathbf N_0^q)&=\sum_{k<q}a_k(\Delta\mathbf W^{(k)},\Delta\mathbf N^{(k)}).
\end{aligned}
\end{equation}
The history stores only the counts generated locally on each query $q_k$, so each local increment enters any later query's initialization at most once. The normalized weights cap the total inherited count mass at one query-equivalent. The inherited counts are converted to $\mathbf P_0^q$ using Eq.~(\ref{eq:pairwise_probability}); the neural model separately learns from the interaction history $D$.

\section{Experiments}\label{sec:experiment}

\begin{table*}[t]
\centering
\caption{Performance across seven benchmarks under two proxy-feedback mechanisms. R.L. is per-query normalized ROUGE-L. W.R. is the final-response win rate against a naive response and is reported as mean$\pm$SEM across users. Dataset results aggregate per-user means from three independent runs; the last row is the macro-average of dataset means. \textbf{Bold} marks the best result and underlining marks the second best.}
\label{tab:results}
\small
\setlength{\tabcolsep}{4pt}
\begin{tabular}{lcccccccc}
\toprule
\multirow{2}{*}{\textbf{Dataset}} &
\multicolumn{2}{c}{\textbf{PersonaAgent}} &
\multicolumn{2}{c}{\textbf{POHF}} &
\multicolumn{2}{c}{\textbf{PDO Sampling}} &
\multicolumn{2}{c}{\textbf{ID-TAP}} \\
\cmidrule(lr){2-3} \cmidrule(lr){4-5} \cmidrule(lr){6-7} \cmidrule(lr){8-9}
 & R.L & W.R. & R.L & W.R. & R.L & W.R. & R.L & W.R. \\
\midrule
LaMP-4     & 0.607 & \underline{70.6$\pm$8.6} & 0.659 & 68.3$\pm$6.7 & \underline{0.660} & 69.4$\pm$7.5 & \textbf{0.781} & \textbf{73.0$\pm$5.5} \\
LaMP-5     & 0.723 & 59.7$\pm$1.8 & \underline{0.822} & 67.5$\pm$1.0 & 0.817 & \underline{68.7$\pm$1.4} & \textbf{0.924} & \textbf{77.2$\pm$0.4} \\
LongLaMP-2 & \underline{0.887} & \textbf{88.8$\pm$1.3} & 0.880 & 79.5$\pm$1.3 & 0.865 & 83.2$\pm$0.7 & \textbf{0.939} & \underline{84.7$\pm$2.0} \\
LongLaMP-3 & 0.882 & \underline{82.3$\pm$3.0} & \underline{0.914} & 80.3$\pm$2.8 & 0.912 & 77.2$\pm$3.5 & \textbf{0.964} & \textbf{85.7$\pm$2.1} \\
LongLaMP-4 & 0.864 & \underline{85.2$\pm$1.3} & 0.868 & 84.9$\pm$1.6 & \underline{0.872} & 84.0$\pm$0.7 & \textbf{0.962} & \textbf{87.9$\pm$0.2} \\
PrefEval   & 0.774 & 59.2$\pm$4.6 & \underline{0.943} & 81.7$\pm$5.8 & 0.886 & \underline{83.3$\pm$3.0} & \textbf{0.956} & \textbf{85.8$\pm$1.7} \\
UltraChat  & 0.369 & 86.7$\pm$2.2 & \underline{0.875} & \underline{90.0$\pm$1.7} & 0.845 & 89.2$\pm$2.9 & \textbf{0.907} & \textbf{90.8$\pm$1.7} \\
\midrule
\textbf{Average}    & 0.729 & 76.1 & \underline{0.852} & 78.9 & 0.837 & \underline{79.3} & \textbf{0.919} & \textbf{83.6} \\
\bottomrule
\end{tabular}
\end{table*}

\subsection{Experimental Setup}
\paragraph{Configuration.}
We adopt \textit{DeepSeek-V3.2}~\citep{deepseekai2025deepseekv32} as the backbone LLM for the agent and \textit{Qwen3-Embedding-4B}~\citep{zhang2025qwen3embedding} as our embedding model. The main experiments use $K=400$ candidate personas per user. For each user, every experimental configuration is run independently three times, and the arithmetic mean of those runs is treated as that user's result. Dataset-level means, standard deviations, and standard errors are computed across these per-user results. Geometric statistics are computed for each user's persona pool before aggregation. $N_{\mathrm{eval}}$ denotes query-level evaluation instances, whereas $n_{\mathrm{user}}$ denotes sampled users in the auxiliary sensitivity and runtime analyses.

\paragraph{Datasets.}
Following \textit{PersonaAgent}, we evaluate on \textbf{LaMP}~\citep{salemi2024lamp}. We also include \textbf{LongLaMP}~\citep{kumar2024longlamp}, \textbf{PrefEval}~\citep{zhao2025llms}, and \textbf{UltraChat}~\citep{ding2023enhancing}, covering short- and long-form personalized generation and conversational settings. Dataset details and statistics are provided in Appendix~\ref{app:experiment_details}.

\begin{figure}[t]
    \centering
    \includegraphics[width=0.49\linewidth]{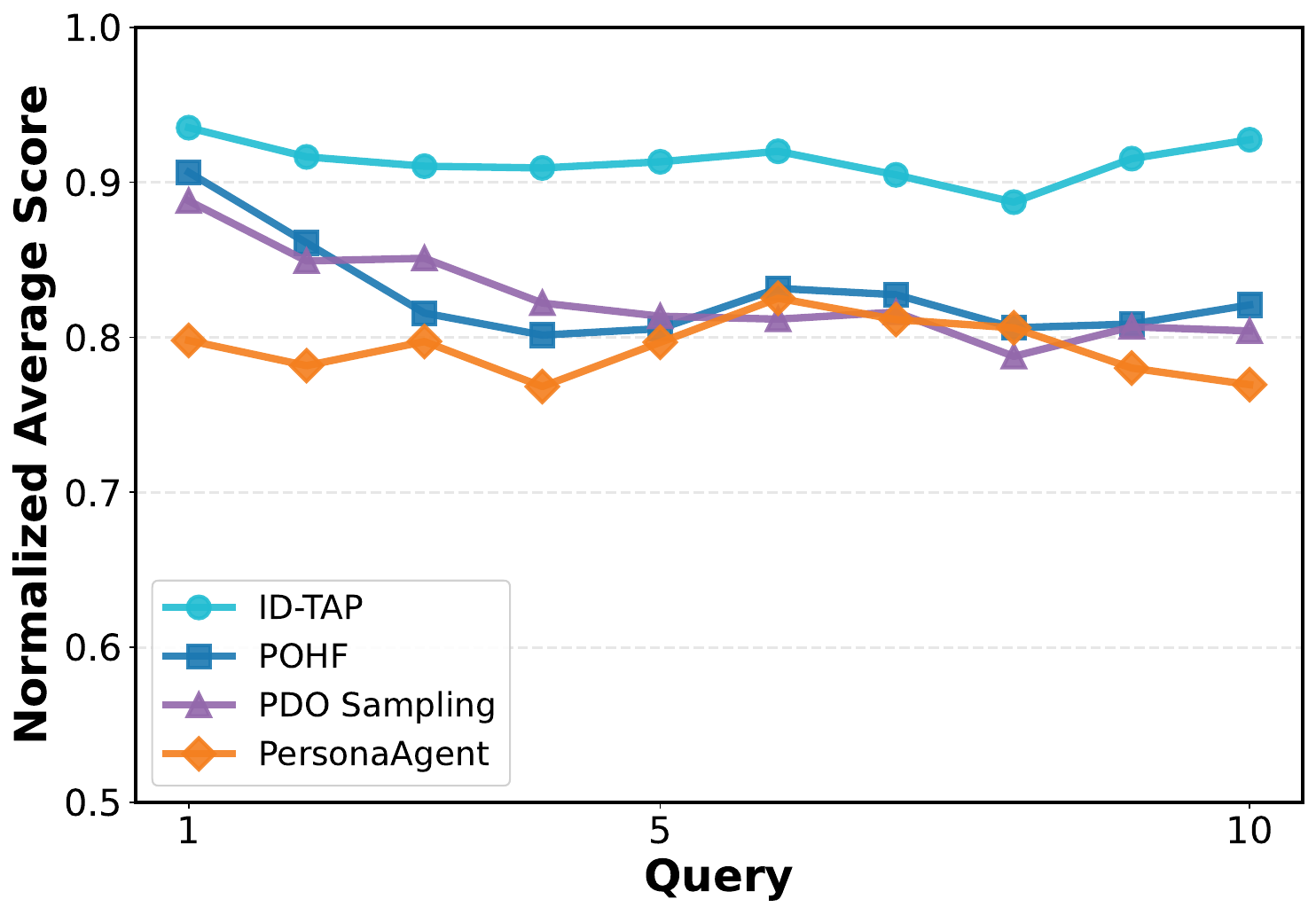}\hfill
    \includegraphics[width=0.49\linewidth]{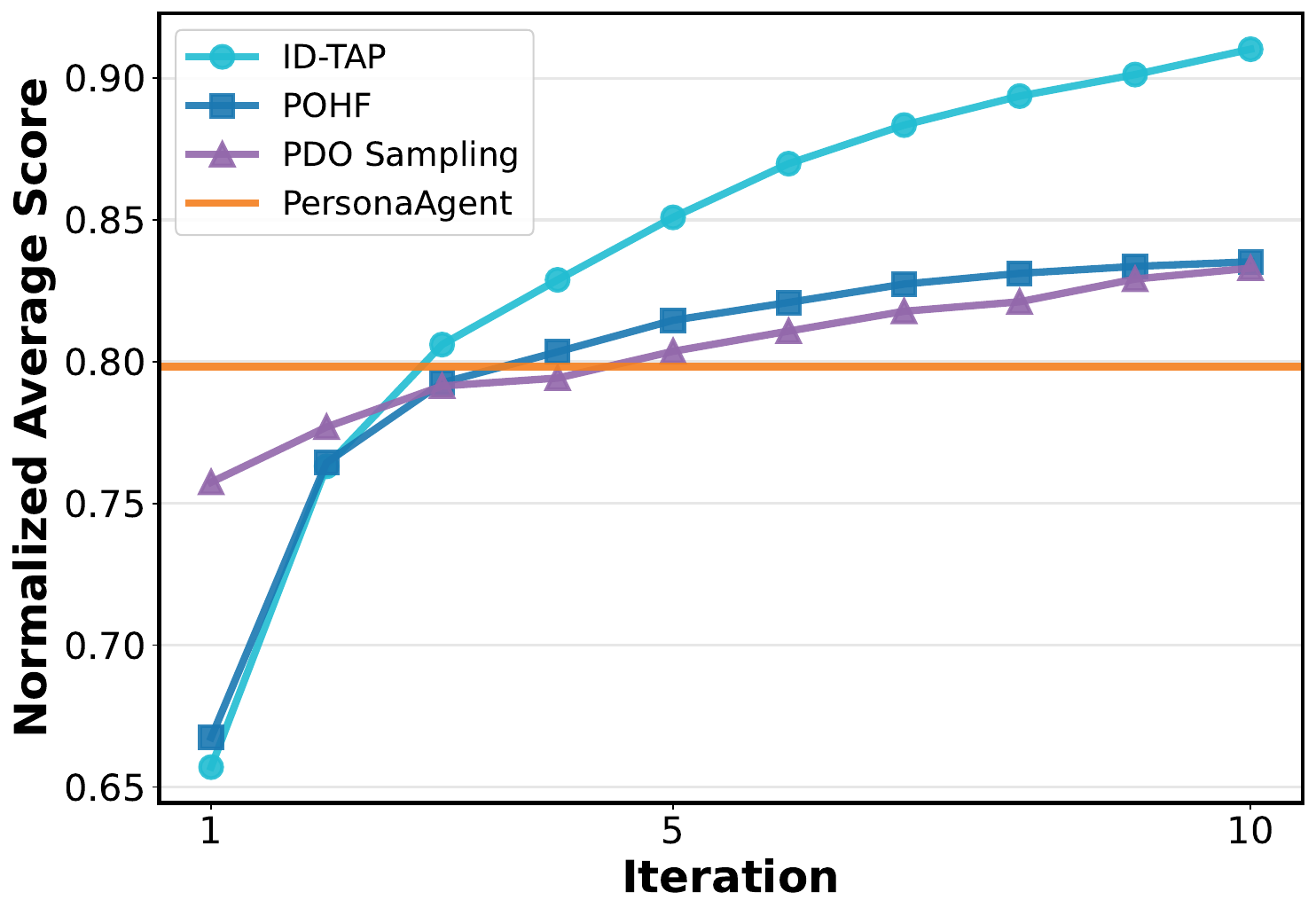}
    \caption{Performance averaged over LaMP-4, LaMP-5, LongLaMP-2, LongLaMP-3, and LongLaMP-4. Left: query-wise normalized ROUGE-L. Right: optimization over ten simulated-preference rounds.}
    \label{fig:lamp_compact_analysis}
\end{figure}

\subsection{Baselines and Shared Evaluation Protocol}
We compare ID-TAP against a generative personalization baseline and two online preference-learning baselines.

\textbf{Generative Baseline.}
\textit{PersonaAgent}~\citep{zhang2025personaagent} performs personalization by synthesizing user-specific information into prompts.

\textbf{Online Learning Baselines.}
Since our setting involves iterative preference learning, we compare against two dueling-bandit acquisition rules: \textit{POHF}, which uses a neural UCB-style rule, and the \textit{PDO sampling component}, i.e., Double Thompson Sampling (D-TS)~\citep{wu2016double,wu2025llm}. Full PDO combines sampling with prompt mutation and pruning. We instantiate its D-TS sampling component on the shared fixed persona pool, yielding a matched comparison of pair selection under the same search space and feedback budget.

\textbf{Shared Evaluation Protocol.} PersonaAgent follows its reported configuration. All bandit methods share the same candidate personas, response generator, binary-feedback interface, and ten-round budget for each query. Within this common protocol, PDO Sampling represents its acquisition component and ID-TAP represents the proposed acquisition design.

\subsection{Preference Feedback Mechanisms}
\label{sec:preference_feedback}
ID-TAP's online update consumes one binary outcome for each displayed response pair. In deployment, the user supplies this outcome after viewing the pair. For repeatable offline experiments, the held-out target instantiates a \emph{proxy user} that returns the same type of outcome after each algorithm-selected comparison. Following POHF's offline interaction protocol~\citep{lin2024prompt}, persona construction, response generation, and arm selection use the history, query, and observed binary outcomes; after pair selection, the proxy user uses the held-out target to produce the next outcome. This protocol reproduces test-time information flow while standardizing the preference source across methods for $T=10$ rounds.

For final W.R. evaluation, each method's final response is paired with a \emph{naive response}: an answer generated by the same backbone directly from the query and user history or context, without persona selection or test-time preference optimization. The pair order is randomized before evaluation, and W.R. is the percentage of comparisons in which the method response is preferred to the naive response.

\paragraph{ROUGE-L-Based Simulation.}
For each candidate response $r_{t,i}^q$, the proxy user computes the reference-grounded utility $s_{t,i}^q=\operatorname{ROUGE\text{-}L}(r_{t,i}^q,r_{gt}^q)$. Following the offline preference-simulation protocol of POHF~\citep{lin2024prompt}, we sample the binary outcome using the Bradley--Terry--Luce model:
\begin{equation}
P(r_{t,1}^q\succ r_{t,2}^q)
=\frac{\exp(s_{t,1}^q)}{\exp(s_{t,1}^q)+\exp(s_{t,2}^q)}.
\end{equation}

\paragraph{Reference-Grounded Holistic Preference Proxy.}
An LLM judge receives the query, a user-specific reference, and an anonymized response pair. For every comparison, the two responses are randomly assigned to the Response~1 and Response~2 positions, and the judge's selection is mapped back to the corresponding personas. The judge selects the response that better realizes the intended output as a whole, jointly considering task fulfillment, semantic intent, organization, coherence, and the tone and style expressed by the reference. The reference supplies a task- and user-conditioned example, and the judge balances semantic realization with organization and style. We refer to this signal as a reference-grounded holistic preference proxy. In both settings, the optimizer's adaptation signal is the resulting binary comparison, while the proxy user retains the reference and its derived score.

\subsection{Main Results and Analysis}
\label{sec:results}

\paragraph{Overall Performance.} 
Table~\ref{tab:results} summarizes final performance across the seven datasets. Under ROUGE-L-based preference simulation, R.L. values are normalized per query by the maximum value among the evaluated methods; under LLM-based preference simulation, W.R. measures how often the method's final response is preferred to the naive response and is reported as mean$\pm$SEM across users. ID-TAP is numerically best in 13 of the 14 dataset--metric cells and has the strongest average in both columns; the exception is LongLaMP-2 W.R., where PersonaAgent is higher. The raw ROUGE-L table yields the same first-place ordering for ID-TAP on all seven datasets, with a cross-dataset average of 0.2228 versus 0.2068 for the strongest baseline. Appendix~\ref{app:raw_rougel} reports unnormalized ROUGE-L as mean$\pm$SEM together with evaluation-set sizes.

\paragraph{Query-Level Performance on Generation Tasks.}
The left panel of Figure~\ref{fig:lamp_compact_analysis} shows query-wise ROUGE-L averaged over the five generation tasks LaMP-4, LaMP-5, LongLaMP-2, LongLaMP-3, and LongLaMP-4. Separate plots for every task, including PrefEval and UltraChat, are provided in Appendix~\ref{app:performance_comparison}.

\paragraph{Feedback-Round Efficiency and Dynamics.} 
We analyze the optimization dynamics under ROUGE-L-based preference simulation in the right panel of Figure~\ref{fig:lamp_compact_analysis}. The plot illustrates iteration-level performance averaged across the five LaMP and LongLaMP generation tasks shown in the figure. ID-TAP exceeds the compared baselines after approximately \textbf{4 simulated-feedback rounds} on this aggregate curve, indicating faster optimization under the shared evaluation protocol. Detailed iteration-level trajectories for all seven tasks, including PrefEval and UltraChat, are provided in Appendix~\ref{app:iteration_trajectories}.

\subsection{Matched Component Ablations}
\label{sec:matched_challenger_ablation}

All variants independently produce their own test-time trajectories under the
same persona pool, query order, comparison budget, model configuration, and
evaluation protocol. \emph{Random Challenger} retains the neural champion,
both updates, cross-query states, and neural final selector, changing only the
challenger to a uniform non-champion draw. \emph{ID-TAP-NH} reinitializes the
neural model and neutralizes the BT statistics for every query, whereas
\emph{ID-TAP-NBT} retains neural history but neutralizes BT statistics; hence
NH versus NBT isolates neural-history transfer and NBT versus Full isolates the
BT warm start. \emph{Random Response} and \emph{Random Acquisition + Neural
Final} are additional diagnostics of random data collection and output
selection.

\begin{table}[H]
\centering
\caption{Normalized ROUGE-L for matched component ablations, aggregated from per-user means over three independent runs. NH
removes neural and BT history; NBT retains neural history but removes BT
transfer.}
\label{tab:ablation}
\footnotesize
\setlength{\tabcolsep}{3pt}
\begin{tabular}{lccc}
\toprule
\textbf{Method} & \textbf{LongLaMP-3} & \textbf{LongLaMP-4} & \textbf{PrefEval} \\
\midrule
Random Response                       & 0.601 & 0.497 & 0.448 \\
Random Acquisition + Neural Final     & 0.615 & 0.516 & 0.507 \\
ID-TAP-NH                             & 0.764 & 0.647 & 0.629 \\
ID-TAP-NBT                            & 0.774 & 0.657 & 0.629 \\
Random Challenger                     & 0.778 & 0.644 & 0.618 \\
\midrule
\textbf{Full ID-TAP}                  & \textbf{0.783} & \textbf{0.664} & \textbf{0.629} \\
\bottomrule
\end{tabular}
\end{table}

Full ID-TAP exceeds Random Challenger by 0.005, 0.020, and 0.011, supporting a
modest contribution from information-directed challenger selection. Along the
history progression NH$\rightarrow$NBT$\rightarrow$Full, neural history adds
0.010 on both LongLaMP tasks, and the BT warm start adds a further 0.009 and
0.007; PrefEval remains unchanged. Thus, both history mechanisms contribute on
the sequential LongLaMP tasks. The wider gaps from the random-acquisition
controls reflect champion-centered collection and learned final selection.

\paragraph{Persona-Construction and Density Analysis.}
Keyword-conditioned construction produces denser persona spaces and lower challenger exploration for POHF and PDO Sampling across the four evaluated datasets, while ID-TAP's exploration rate changes by at most 1.5 percentage points. Appendix~\ref{app:persona_density_analysis} reports the construction, exploration, performance, and sensitivity results.

\paragraph{Robustness to Preference and Evaluation Models.}
ID-TAP is numerically highest across the reported matched-model, cross-model, and BERTScore configurations, providing complementary evidence across automated preference sources, evaluators, and metrics. Appendix~\ref{app:robustness} provides the complete tables and protocol.

\paragraph{Runtime.}
Here, $K$ denotes the number of candidate personas in each user's persona space. With $K=100$ and $n_{\mathrm{user}}=40$ sampled users per dataset, ID-TAP's algorithmic update requires 0.065--0.571 seconds per round, compared with 0.95--3.15 seconds of end-to-end wall-clock latency. The algorithmic component accounts for approximately 7--22\% of total latency. Full timing statistics and the measurement protocol are provided in Appendix~\ref{app:runtime}.

\section{Related Work}\label{sec:related}
\paragraph{Personalized LLM Agents.}
Personalized LLMs condition outputs on individual preferences, traits, or context~\citep{li2024personal,zhu2024personality}. Retrieval-based methods expose relevant user history at inference time~\citep{richardson2023integrating,zhong2024memorybank}, while memory and prompt-based approaches maintain or synthesize user-specific context~\citep{zhang2025personaagent,liu2024test}. LaMP standardizes evaluation across personalized language tasks~\citep{salemi2024lamp}. Recent work increasingly studies lightweight adaptation at inference time~\citep{li2024personal,qu2025t}; ID-TAP contributes a pairwise-feedback method for selecting among fixed persona prompts.

\paragraph{Feedback Mechanisms and Test-Time Adaptation.} Aligning LLM behaviors with user intent relies heavily on feedback-driven optimization. The canonical paradigm, Reinforcement Learning from Human Feedback (RLHF) \citep{christiano2017deep,ziegler2019fine}, typically employs supervised fine-tuning, reward modeling \citep{stiennon2020learning}, and policy optimization via PPO \citep{schulman2017proximal,ouyang2022training}. Direct Preference Optimization (DPO) \citep{rafailov2023direct} and RLAIF \citep{bai2022constitutional} streamline this process through direct preference learning or AI supervision. Test-Time Adaptation (TTA) further enables online refinement through iterative feedback loops \citep{madaan2023self,shinn2023reflexion}; reasoning systems use self-correction or tree search \citep{yao2023tree,ouyang2025reasoningbank}, while personalized prompt- and decoding-level methods adapt behavior with fixed backbone LLM weights \citep{lin2024prompt,qu2025t}.

\paragraph{Exploration in Dueling Bandits.} Dueling bandits learn from pairwise preferences~\citep{yue2012k}, and contextual variants allow preferences to depend on the query~\citep{dudik2015contextual,bengs2022stochastic}. Relevant analyses cover structured stochastic preferences and neural utility models~\citep{bengs2022stochastic,verma2025neuralduelingbanditspreferencebased}; confidence-based linear bandit methods provide additional background for contextual exploration~\citep{abbasi2011improved}. Practical strategies include Thompson sampling~\citep{daniel2018tutorial}, Double Thompson Sampling (D-TS)~\citep{wu2016double}, and posterior sampling for LLM exploration~\citep{dwaracherla2024efficientexplorationllms}. For LLM prompt optimization, POHF~\citep{lin2024prompt} uses neural UCB-style acquisition, and PDO~\citep{wu2025llm} combines D-TS with top-performer-guided mutation and pruning. Our PDO Sampling baseline instantiates D-TS in the common fixed persona pool and feedback budget. T-POP~\citep{qu2025t} operates during token-level decoding with white-box model access and addresses a complementary setting.

IDS balances estimated utility gaps against information gain~\citep{russo2018learning} and has been studied in linear partial monitoring and dueling-feedback settings~\citep{kirschner2020information,kirschner2021bias,pasztor2024bandits}. These works provide general information-directed objectives or analyses under structured models. ID-TAP builds on these foundations and focuses on the systems question posed by black-box LLM personalization: scoring a large, LLM-generated persona pool, separating query-conditioned neural utility learning from lightweight hypothetical BT updates, and transferring comparison evidence across related queries. The resulting contribution is a scalable, cross-query acquisition architecture for persona-level test-time adaptation.

\section{Conclusion}
\label{sec:conclusion}

We presented \textbf{ID-TAP}, an IDS-inspired framework for test-time persona selection. It combines query-conditioned neural utility gaps, a lightweight BT entropy-reduction proxy, and similarity-weighted warm starts across related queries. Across seven user-conditioned tasks with automated proxy feedback, ID-TAP achieves the highest aggregate mean among the evaluated baselines under a fixed comparison budget. Future work will evaluate the interface with end users and develop finite-sample analysis of the neural/BT procedure.

\subsection*{AI use statement}
In preparing this manuscript, generative AI tools were used to assist with
\LaTeX{} formatting, scientific-figure design and implementation, and editorial
review. These tools did not generate or alter the reported experimental
measurements. The authors reviewed all AI-assisted changes and take
responsibility for the final text, claims, and artifacts.

\subsection*{Reproducibility statement}
The main paper specifies the problem formulation, ID-TAP procedure, and
evaluation protocol. The appendix provides the BT-MLE entropy-proxy
implementation, complete prompts, data splits, hyperparameters, persona-space
construction analyses, robustness experiments, and timing protocol needed to
interpret and reproduce the reported results. The public code artifact is
available at \url{https://github.com/s1412/ID-TAP}.

\bibliographystyle{iclr2027_conference}
\bibliography{example_paper}

\clearpage
\appendix
\section{Implementation Details}
\label{app:bt_mi_impl}
\subsection{Cross-Query BT Warm Start}
Algorithm~\ref{algo:inheritance} gives the complete warm-start procedure corresponding to the normalized transfer rule in Section~\ref{sec:method}. Its history contains query-local count increments, and the normalized weights provide at most one query-equivalent of inherited mass.

\begin{algorithm}[H]
    \caption{Cross-Query BT Probability Inheritance}
    \label{algo:inheritance}
    \begin{algorithmic}[1]
    \STATE \textbf{Input:} Current embedding $\mathbf e_q$; local history $\mathcal H=\{(q_k,\mathbf e_{q_k},\Delta\mathbf W^{(k)},\Delta\mathbf N^{(k)})\}_{k<q}$
    \STATE \textbf{Output:} Initialized $(\mathbf{W}^{q},\mathbf{N}^{q},\mathbf{P}^{q})$
    \IF{$\mathcal{H} = \emptyset$}
        \STATE \textbf{return} $\mathbf{W}^{q},\mathbf{N}^{q}\leftarrow\mathbf{0}$, $\mathbf{P}^{q}\leftarrow\mathbf{0.5}$
    \ENDIF
    \STATE $b_k\leftarrow\sigma\!\left(\beta_{\mathrm{sim}}(\text{CosineSim}(\mathbf e_q,\mathbf e_{q_k})-\mu_{\mathrm{sim}})\right)$ for every $k<q$
    \STATE $Z_q\leftarrow\max\{1,\sum_{k<q}b_k\}$; $\mathbf W^q,\mathbf N^q\leftarrow\mathbf0$
    \FOR{each $(q_k,\mathbf e_{q_k},\Delta\mathbf W^{(k)},\Delta\mathbf N^{(k)})\in\mathcal H$}
        \STATE $a_k\leftarrow b_k/Z_q$
        \STATE $\mathbf W^q\leftarrow\mathbf W^q+a_k\Delta\mathbf W^{(k)}$, \quad $\mathbf N^q\leftarrow\mathbf N^q+a_k\Delta\mathbf N^{(k)}$
    \ENDFOR
    \STATE $\mathbf{P}^{q}_{ij} \leftarrow \frac{\mathbf{W}^{q}_{ij} + \alpha}{\mathbf{N}^{q}_{ij} + 2\alpha}$
    \STATE \textbf{return} $(\mathbf{W}^{q},\mathbf{N}^{q},\mathbf{P}^{q})$
    \end{algorithmic}
\end{algorithm}

\subsection{Maintained Statistics and Smoothed Pairwise Probabilities}
For each persona pair $(p_i,p_j)$, we maintain (i) directed win counts $w_{ij}$ and (ii) symmetric comparison counts $n_{ij}=n_{ji}$.
After each observed duel, counts are updated as:
\begin{equation}
w_{ij} \leftarrow w_{ij} + \mathbb{I}\{p_i \succ p_j\}, \quad
w_{ji} \leftarrow w_{ji} + \mathbb{I}\{p_j \succ p_i\}, \quad
n_{ij} \leftarrow n_{ij}+1,\; n_{ji} \leftarrow n_{ji}+1.
\end{equation}
Observed pairs receive the symmetric Beta-smoothed estimate
\begin{equation}
P_{ij}=\frac{w_{ij}+\alpha}{w_{ij}+w_{ji}+2\alpha},
\label{eq:pairwise_probability}
\end{equation}
where $\alpha=0.7$. Unobserved pairs use $P_{ij}=0.5$, preserving neutral pairwise evidence until a direct or inherited comparison becomes available.

\subsection{BT-MLE Strength Estimation via MM Iterations}
Given the current count matrices, we estimate BT strengths $\hat{\theta}$ by an MM-style iterative solver over positive parameters $\pi_i=\exp(\theta_i)$.
Let
\begin{equation}
W_i=\sum_{j\neq i} \tilde{w}_{ij}, \quad
\tilde{w}_{ij}=w_{ij}+\alpha\,\mathbb{I}\{n_{ij}>0\},
\end{equation}
and
\begin{equation}
\tilde{n}_{ij}=n_{ij}+2\alpha\,\mathbb{I}\{n_{ij}>0\}.
\end{equation}
Each MM iteration updates
\begin{equation}
\pi_i \leftarrow
\frac{W_i}{\sum_{j\neq i}\frac{\tilde{n}_{ij}}{\pi_i+\pi_j}},
\label{eq:app_mm_update}
\end{equation}
for candidates with positive incident comparison mass. Candidates awaiting their first observed or inherited comparison retain the common neutral initialization. The updated strengths are then normalized (sum rescaling) and converted as $\hat{\theta}_i=\log(\pi_i+\epsilon)$.
For numerical stability we use $\epsilon=10^{-8}$ and clip probabilities to $[\epsilon,1-\epsilon]$.

\subsection{Normalized Strength Distribution and Entropy}
From $\hat{\theta}$, we define the normalized plug-in score distribution $\tilde\pi_t(k)=\operatorname{softmax}(\hat\theta_t/\tau)_k$ for the fixed query $q$ and use it to rank hypothetical updates. Unobserved pairs enter the solver with neutral pairwise evidence. Candidates awaiting their first comparison receive neutral initial strengths, and one softmax over all $K$ strengths ensures $\sum_k\tilde\pi_t(k)=1$.
We then compute
\begin{equation}
H(\tilde\pi_t) = -\sum_{k=1}^K \tilde\pi_t(k)\log \tilde\pi_t(k),
\end{equation}
where we use the natural logarithm throughout.

\subsection{Outcome Simulation and Entropy-Reduction Proxy}
For a champion--challenger pair $(c,d)$, outcome probability under BT is
\begin{equation}
P(y=1\mid c,d)=
\begin{cases}
\sigma(\hat{\theta}_c-\hat{\theta}_d), & n_{cd}>0,\\
0.5, & \text{otherwise},
\end{cases}
\label{eq:app_outcome_prob}
\end{equation}
where $y=1$ means champion wins.

To evaluate the proxy, we simulate both hypothetical outcomes $y\in\{0,1\}$:
\begin{enumerate}
    \item update $(w,n)$ with the hypothetical result;
    \item re-estimate BT strengths by Eq.~(\ref{eq:app_mm_update});
    \item recompute the normalized strength distribution $\tilde\pi_t^{(y)}$ and its entropy $H(\tilde\pi_t^{(y)})$.
\end{enumerate}
Thus, the expected post-update entropy is
\begin{equation}
\begin{aligned}
\mathbb{E}[H_{\mathrm{after}}]={}&P(y=1)H(\tilde\pi_t^{(1)})\\
&+(1-P(y=1))H(\tilde\pi_t^{(0)}),
\end{aligned}
\end{equation}
and $\tilde g_t(c,d)=\max\{0,H(\tilde\pi_t)-\mathbb{E}[H_{\mathrm{after}}]\}$ is the plug-in entropy-reduction proxy used by the acquisition rule. Clipping at zero removes tiny negative values caused by floating-point error.

\subsection{Reproducible Procedure (Pseudo-code)}
\begin{algorithm}[H]
\caption{BT-MLE Entropy-Reduction Proxy for a Candidate Pair}
\label{alg:app_bt_mi}
\begin{algorithmic}[1]
\STATE \textbf{Input:} champion $c$, challenger $d$, counts $(w,n)$, smoothing $\alpha$
\STATE Estimate $\hat{\theta}$ from $(w,n)$ via MM updates (Eq.~\ref{eq:app_mm_update})
\STATE Compute $\tilde\pi_t=\operatorname{softmax}(\hat\theta/\tau)$ and $H_{\text{before}}=H(\tilde\pi_t)$
\STATE Compute $p_{\text{win}}=P(y=1\mid c,d)$ using Eq.~(\ref{eq:app_outcome_prob})
\FOR{$y\in\{1,0\}$}
    \STATE Simulate outcome $y$ and obtain hypothetical counts $(w^{(y)},n^{(y)})$
    \STATE Re-estimate $\hat{\theta}^{(y)}$ from $(w^{(y)},n^{(y)})$
    \STATE Compute $\tilde\pi_t^{(y)}$ and entropy $H_y=H(\tilde\pi_t^{(y)})$
\ENDFOR
\STATE $\mathbb{E}[H_{\text{after}}] \leftarrow p_{\text{win}}H_1 + (1-p_{\text{win}})H_0$
\STATE \textbf{return} $\tilde g_t(c,d)=\max\{0, H_{\text{before}}-\mathbb{E}[H_{\text{after}}]\}$
\end{algorithmic}
\end{algorithm}

\paragraph{Complexity.}
For each candidate challenger, the proxy performs two hypothetical BT re-estimations, one for each outcome. With a fixed number of MM iterations, an acquisition round therefore requires $2(K-1)$ BT re-estimations.

\section{Experiment Details}
\label{app:experiment_details}
\subsection{Dataset} 

\paragraph{LaMP}
LaMP evaluates personalization from limited user history across several language tasks. We use its two generative tasks:

\begin{itemize}
    \item \textbf{LaMP-4 (Personalized News Headline Generation):} Given an article, generate a headline consistent with the user's historical headline style.
    
    \item \textbf{LaMP-5 (Personalized Scholarly Title Generation):} Given an abstract, generate a paper title consistent with the user's publication history.
\end{itemize}

\paragraph{LongLaMP}
LongLaMP extends personalized evaluation to longer-form generation. We use three tasks:

\begin{itemize}
    \item \textbf{LongLaMP-2 (Personalized Abstract Generation):} Generate an abstract from a title and keywords in the user's academic style.
    
    \item \textbf{LongLaMP-3 (Personalized Product Review Generation):} Generate a review from a product description, rating, and review summary in the user's style.
    
    \item \textbf{LongLaMP-4 (Personalized Topic Generation):} Expand a topic summary into a Reddit post consistent with the user's history.
\end{itemize}

\paragraph{PrefEval}
PrefEval is a multi-turn dialogue dataset designed for evaluating preference following. We adapt it by treating the dialogue history as persona context $\mathcal{P}(\mathcal{C})$. The final user query is used as the input $q$, and the corresponding assistant response serves as the held-out reference. The agent is expected to answer $q$ while remaining consistent with the inferred user persona.

\paragraph{UltraChat}
UltraChat contains multi-turn human--AI conversations. We condition on the dialogue history and predict the \textbf{user's next query} $Q$, treating next-turn prediction as a user-conditioned generation task distinct from assistant-response generation.

{

\subsection{Prompt Templates}
\label{app:prompts}

This section presents all prompt templates used in ID-TAP, including persona construction, response generation, and evaluation.

\paragraph{Persona Summarization Prompt.}
The persona space $\mathcal{P}(\mathcal{C})$ is constructed offline from the user's historical context $\mathcal{C}$. We first generate a base persona summary using the following prompt:

\begin{tcolorbox}[title={\textbf{\small Persona Summarization Prompt}}, boxrule=1pt, arc=0mm, colback=black!5!white, colframe=black!75!white, breakable, before skip=10pt, after skip=10pt, pad at break=2mm, parbox=false]
\begin{lstlisting}
=== SYSTEM ===
You are an advanced profile analysis and summarization model. Your goal is to generate a coherent, SPECIFIC, and psychologically insightful user persona summary.

=== USER ===
Based on the user's historical data provided below, generate an informative and coherent user persona summary.

### Task Instruction: {instruction}

### User History Data: {history_context}

### Instructions:
1. CORE ANCHORING: Identify 2-3 SPECIFIC, CONCRETE details from the history
2. BEHAVIORAL PATTERNS: Focus on personality traits, interests, tone, communication style
3. SPECIFICITY OVER ABSTRACTION: Use CONCRETE language, avoid vague terms

Please provide the final user persona summary (one paragraph, 150-300 words).
\end{lstlisting}
\end{tcolorbox}

\paragraph{Persona Rephrasing Prompt.}
To construct a diverse persona space of size $K$, we iteratively rephrase the base summary. Each variant $p_i$ is generated by rephrasing the previous variant $p_{i-1}$ with bounded variation. To prevent semantic drift, we periodically reset to the original history context every $R=15$ iterations.

\begin{tcolorbox}[title={\textbf{\small Persona Rephrasing Prompt}}, boxrule=1pt, arc=0mm, colback=black!5!white, colframe=black!75!white, breakable, before skip=10pt, after skip=10pt, pad at break=2mm, parbox=false]
\begin{lstlisting}
=== SYSTEM ===
You are an advanced language model specialized in balanced rephrasing of personal profiles. Your task is to rewrite a user persona summary with a SLIGHTLY DIFFERENT perspective or emphasis.

=== USER ===
Rewrite the following user persona summary with a SLIGHTLY DIFFERENT perspective:

{previous_persona}

### STRICT Constraints:
- PRESERVE CORE ANCHORS: Keep 2-3 specific concrete details from the original
- BOUNDED VARIATION: Change emphasis or perspective, but the person should still be recognizable
- NO EXTREME SHIFTS: Do NOT flip personality traits dramatically
\end{lstlisting}
\end{tcolorbox}

\paragraph{Response Generation Prompt.}
The following prompt template is used to generate persona-conditioned responses:

\begin{tcolorbox}[title={\textbf{\small Response Generation Prompt}}, boxrule=1pt, arc=0mm, colback=black!5!white, colframe=black!75!white, breakable, before skip=10pt, after skip=10pt, pad at break=2mm, parbox=false]
\begin{lstlisting}
### Task Instruction: {instruction}

### Current Task Details: {query}

### Personality and Style Description: {persona_summary}

### Important Words from history: {keywords}

### History context: {history_entries}

### Output Requirement:
Complete the task according to the Task Instruction and Current Task Details. Prioritize task accuracy and relevance while using the Personality and Style Description to guide tone. Provide only the required response below:
\end{lstlisting}
\end{tcolorbox}

\paragraph{LLM-as-Judge Evaluation Prompt.}
For pairwise preference evaluation, we employ an LLM-as-Judge approach. During online adaptation, the pair contains the two persona-conditioned responses selected for that round. For final W.R. evaluation, the pair contains the method's final response and the naive response generated directly from the query and user history or context, without persona selection or test-time preference optimization. Before each judge call, the two responses are randomly assigned to the Response~1 and Response~2 positions; the returned label is then mapped back to the corresponding source.

\begin{tcolorbox}[title={\textbf{\small LLM-as-Judge System Role}}, boxrule=1pt, arc=0mm, colback=black!5!white, colframe=black!75!white, breakable, before skip=10pt, after skip=10pt, pad at break=2mm, parbox=false]
\begin{lstlisting}
You are an expert evaluator comparing two candidate responses for the same personalized task. The user-specific reference exemplifies the intended content and realization. Decide which response better fulfills the intended output as a whole. Jointly consider task fulfillment, semantic fidelity, organization and coherence, and the tone and writing style expressed by the reference, treating surface phrase overlap as secondary to the overall realization. Respond with ONLY a single digit: 1 if Response 1 is better, or 0 if Response 2 is better.
\end{lstlisting}
\end{tcolorbox}

\begin{tcolorbox}[title={\textbf{\small Comparison Prompt}}, boxrule=1pt, arc=0mm, colback=black!5!white, colframe=black!75!white, breakable, before skip=10pt, after skip=10pt, pad at break=2mm, parbox=false]
\begin{lstlisting}
Current Query: {query}

User-Specific Reference: {ground_truth}

Response 1: {response_1}

Response 2: {response_2}

Which response better realizes the intended output overall? Reply with only 1 (Response 1 is better) or 0 (Response 2 is better). If they are equivalent overall, return 0 by default. Please provide the final result (1 or 0 only) below:
\end{lstlisting}
\end{tcolorbox}

\subsection{Data Splitting for LaMP Datasets}
\label{app:data_split}

For LaMP-based datasets (LaMP-4, LaMP-5, LongLaMP-2, LongLaMP-3, LongLaMP-4), each user's historical profile is split into two parts: (1) \textbf{history context} used for persona generation, and (2) \textbf{input/output pairs} used as test queries with held-out reference responses.

\paragraph{Splitting Rules.}
Let $n$ denote the number of profile entries for a user. The splitting follows a threshold-based strategy:

\begin{itemize}
    \item If $n < T_{\text{small}}$: The first $H_{\text{small}}$ entries form the history context; the remaining entries become input/output pairs.
    \item If $T_{\text{small}} \leq n < T_{\text{large}}$: The first $H_{\text{large}}$ entries form the history context; the remaining entries become input/output pairs.
    \item If $n \geq T_{\text{large}}$: The first $H_{\text{large}}$ entries form the history context; entries from position $H_{\text{large}}+1$ to $H_{\text{large}} + (N_{\text{IO}} - 1)$ become input/output pairs (capped at $N_{\text{IO}} - 1$ additional pairs).
\end{itemize}

\begin{table}[h]
\centering
\caption{Data Splitting Parameters}
\label{tab:data_split}
\begin{tabular}{lcc}
\toprule
\textbf{Parameter} & \textbf{Symbol} & \textbf{Value} \\
\midrule
Small profile threshold & $T_{\text{small}}$ & 10 \\
Large profile threshold & $T_{\text{large}}$ & 20 \\
History context count (small) & $H_{\text{small}}$ & 5 \\
History context count (large) & $H_{\text{large}}$ & 10 \\
Total I/O pairs limit & $N_{\text{IO}}$ & 10 \\
\bottomrule
\end{tabular}
\end{table}

\paragraph{Query and Output Extraction.}
For each dataset, queries and held-out reference outputs are extracted from profile entries as follows:
\begin{itemize}
    \item \textbf{LaMP-4} (News Headline): Query = article text; Output = headline title
    \item \textbf{LaMP-5} (Paper Title): Query = abstract; Output = paper title
    \item \textbf{LongLaMP-2} (Abstract Generation): Query = paper title; Output = abstract
    \item \textbf{LongLaMP-3} (Product Review): Query = product description + rating; Output = review text
    \item \textbf{LongLaMP-4} (Reddit Post): Query = topic summary; Output = post content
\end{itemize}

The history context supports persona summarization and construction of $\mathcal{P}(\mathcal{C})$. Each input defines a test query, while its held-out output conditions the offline proxy user and evaluation. During adaptation, the proxy user retains this reference and ID-TAP receives the resulting binary comparison.

\subsection{Unnormalized ROUGE-L Statistics}
\label{app:raw_rougel}

For each user, raw (unnormalized) ROUGE-L is averaged over three independent runs. Table~\ref{tab:raw_rougel} reports the mean and SEM across these per-user results. $N_{\mathrm{eval}}$ denotes the number of query-level evaluation instances contributing to the user-level aggregates. In our logging notation, \textit{lamp8/lamp9/lamp10} correspond to \textit{LongLaMP-2/LongLaMP-3/LongLaMP-4}.

\begin{table*}[t]
\centering
\caption{Unnormalized ROUGE-L performance before per-query normalization. Results are mean$\pm$SEM across users, where each user's result averages three independent runs. \textbf{Bold} indicates the best result and underlined values indicate the second-best result in each dataset.}
\label{tab:raw_rougel}
\small
\setlength{\tabcolsep}{5pt}
\begin{tabular}{lccccc}
\toprule
\textbf{Dataset} & \textbf{$N_{\mathrm{eval}}$} & \textbf{PersonaAgent} & \textbf{POHF} & \textbf{PDO Samp.} & \textbf{ID-TAP} \\
\midrule
LaMP-4     & 363 & 0.1700$\pm$0.007846 & \underline{0.1790$\pm$0.007499} & 0.1768$\pm$0.007085 & \textbf{0.1973$\pm$0.006559} \\
LaMP-5     & 400 & 0.3924$\pm$0.009460 & \underline{0.4385$\pm$0.008671} & 0.4346$\pm$0.008551 & \textbf{0.4927$\pm$0.008513} \\
LongLaMP-2 & 400 & \underline{0.1729$\pm$0.002141} & 0.1716$\pm$0.002119 & 0.1678$\pm$0.001918 & \textbf{0.1831$\pm$0.002042} \\
LongLaMP-3 & 230 & 0.1322$\pm$0.001532 & \underline{0.1391$\pm$0.001712} & 0.1378$\pm$0.001251 & \textbf{0.1456$\pm$0.001562} \\
LongLaMP-4 & 326 & 0.1217$\pm$0.001576 & \underline{0.1234$\pm$0.001806} & \underline{0.1234$\pm$0.001547} & \textbf{0.1367$\pm$0.001853} \\
PrefEval   & 40  & 0.1700$\pm$0.005887 & \underline{0.2103$\pm$0.006989} & 0.1943$\pm$0.004623 & \textbf{0.2131$\pm$0.007093} \\
UltraChat  & 40  & 0.0758$\pm$0.005492 & \underline{0.1859$\pm$0.009107} & 0.1815$\pm$0.009301 & \textbf{0.1909$\pm$0.008171} \\
\midrule
\textbf{Average} & -- & 0.1764 & \underline{0.2068} & 0.2023 & \textbf{0.2228} \\
\bottomrule
\end{tabular}
\end{table*}

ID-TAP has the highest raw ROUGE-L mean on all seven datasets, with a cross-dataset average of 0.2228 compared with 0.2068 for the strongest baseline. These unnormalized values provide a scale-preserving counterpart to the normalized main results.

\subsection{Persona Space Quality Analysis}
\label{app:persona_space_quality}
 
To characterize the generated persona spaces, Table~\ref{tab:persona_space_quality} reports five geometric metrics for all seven benchmarks. The analysis covers 180 users, with 400 candidate personas per user. Here, $N_{\mathrm{user}}$ is 20 for each LaMP/LongLaMP dataset and 40 for PrefEval and UltraChat.
 
\paragraph{Density \& Spread.}
The mean pairwise cosine similarity $\bar{\mu}$ ranges from 0.657 (LongLaMP-3) to 0.716 (PrefEval), with standard deviations $\bar{\sigma}$ between 0.080 and 0.093. The 10th--90th percentile interval ($p_{10}$--$p_{90}$) spans roughly 0.22 in each dataset (e.g., [0.537, 0.769] for LongLaMP-3). These statistics describe moderately dense but non-saturated embedding spaces.
 
\paragraph{Redundancy.}
The near-duplicate rate (fraction of pairs with cosine similarity $\geq0.95$) stays below 0.20\%, and the effective-arm ratio---the percentage of candidates whose nearest-neighbor similarity is below 0.95---ranges from 94.4\% to 98.2\%. Under this threshold, the pools contain little near-duplicate redundancy while retaining many distinct candidate embeddings.
 
\paragraph{Geometric Structure.}
K-means silhouette scores at $K{=}20$ range from 0.097 (UltraChat) to 0.125 (LaMP-5), indicating weak separation into discrete clusters. Together with the similarity statistics, these scores characterize the candidate pools as dense and weakly clustered; the acquisition experiments measure the corresponding method-level behavior.

\begin{table*}[t]
\centering
\caption{Persona space quality metrics across 7 benchmarks (180 users in total; 400 candidates per user). $N_{\mathrm{user}}$ is the number of sampled users in each dataset. $\bar{\mu}$: mean pairwise cosine similarity; $\bar{\sigma}$: standard deviation; $p_{10}$/$p_{90}$: 10th/90th percentile of the pairwise similarity distribution; Near-dup: fraction of pairs with similarity $\geq 0.95$; Eff.\ Arm: percentage of candidates whose nearest-neighbor similarity $< 0.95$; Silhouette: K-means silhouette coefficient at $K{=}20$.}
\label{tab:persona_space_quality}
\resizebox{\textwidth}{!}{%
\begin{tabular}{l r r r r r r r r}
\toprule
Dataset & $N_{\mathrm{user}}$ & $\bar{\mu}$ & $\bar{\sigma}$ & $p_{10}$ & $p_{90}$
 & \shortstack{Near-dup\\$\geq$0.95 (\%)} & \shortstack{Eff.\ Arm\\(\%)}
 & \shortstack{Silhouette\\($K$=20)} \\
\midrule
LaMP-4     & 20 & 0.672 & 0.090 & 0.553 & 0.779 & 0.041 & 97.6 & 0.104 \\
LaMP-5     & 20 & 0.694 & 0.081 & 0.590 & 0.790 & 0.094 & 95.3 & 0.125 \\
LongLaMP-2     & 20 & 0.701 & 0.080 & 0.599 & 0.794 & 0.092 & 95.0 & 0.121 \\
LongLaMP-3     & 20 & 0.657 & 0.093 & 0.537 & 0.769 & 0.018 & 98.2 & 0.107 \\
LongLaMP-4     & 20 & 0.686 & 0.084 & 0.577 & 0.787 & 0.037 & 97.5 & 0.105 \\
PrefEval   & 40 & 0.716 & 0.083 & 0.607 & 0.814 & 0.190 & 94.4 & 0.098 \\
UltraChat  & 40 & 0.715 & 0.087 & 0.600 & 0.819 & 0.158 & 94.5 & 0.097 \\
\bottomrule
\end{tabular}%
}
\end{table*}

\subsection{Hyperparameters}
\label{app:hyperparams}

This section details all hyperparameters used in ID-TAP, organized into four categories: neural network architecture, training configuration, algorithm-specific parameters, and LLM settings. Table~\ref{tab:all_hyperparams} provides a comprehensive summary.

\paragraph{Neural Network Architecture.}
The utility function $f_\theta(q, p)$ is parameterized by a single-hidden-layer feedforward neural network. The input consists of the concatenation of query embedding $\phi(q) \in \mathbb{R}^{d_{\text{emb}}}$ and persona embedding $\phi(p) \in \mathbb{R}^{d_{\text{emb}}}$, where $d_{\text{emb}} = 1024$ for standard mode and $d_{\text{emb}} = 2048$ for contextual mode. We employ GELU activation for its smooth gradient properties and use a dropout rate of $0.05$ for regularization. Weights are initialized using Kaiming normal initialization to ensure proper gradient flow at initialization.

\paragraph{Training Configuration.}
The neural network is trained using the AdamW optimizer with a learning rate of $5 \times 10^{-4}$ and weight decay of $10^{-3}$. We use a cosine annealing learning rate scheduler that decays the learning rate to 50\% of its initial value over the training epochs. Each update consists of 100 epochs with a batch size of 4, and gradient clipping with a maximum norm of 1.0 is applied to prevent gradient explosion during early training stages when comparison data may be noisy.

\paragraph{ID-TAP Algorithm Parameters.}
The acquisition ratio in Section~\ref{sec:method} uses a numerical floor $\epsilon_g=10^{-8}$. The BT explorer uses smoothing $\alpha=0.7$ and softmax temperature $\tau=4.5$ for the normalized strength distribution. Cross-query transfer uses $\beta_{\mathrm{sim}}=6$ and $\mu_{\mathrm{sim}}=0.8$ in the normalized similarity weights.

\paragraph{LLM Configuration.}
The main configuration uses DeepSeek-V3.2 for response generation and for the LLM-based preference proxy and final evaluation; the ROUGE-L-based simulator follows the scoring rule in Section~\ref{sec:preference_feedback}. Appendix~\ref{app:robustness} varies the preference and evaluation models. Generation uses temperature 0.6, whereas LLM-as-Judge evaluation uses temperature 0.0. The maximum output length is 1024 tokens, with frequency and presence penalties of 0.5.

\begin{table*}[t]
\centering
\caption{Complete hyperparameter configuration for ID-TAP. The table is organized into four sections: neural network architecture, training settings, algorithm-specific parameters, and LLM configuration.}
\label{tab:all_hyperparams}
\small
\begin{tabular}{llcc}
\toprule
\textbf{Category} & \textbf{Parameter} & \textbf{Symbol} & \textbf{Value} \\
\midrule
\multirow{5}{*}{\textit{Neural Network}}
& Input dimension & $d_{\text{in}}$ & $2 \times d_{\text{emb}}$ \\
& Hidden dimension & $d_h$ & 1024 \\
& Number of hidden layers & $L$ & 1 \\
& Activation function & $\sigma(\cdot)$ & GELU \\
& Dropout rate & $p_{\text{drop}}$ & 0.05 \\
\midrule
\multirow{7}{*}{\textit{Training}}
& Learning rate & $\eta$ & $5 \times 10^{-4}$ \\
& Optimizer & -- & AdamW \\
& Weight decay & $\lambda$ & $10^{-3}$ \\
& Batch size & $B$ & 4 \\
& Update epochs & -- & 100 \\
& Gradient clipping & -- & 1.0 \\
& Learning-rate schedule & -- & Cosine to $0.5\eta$ \\
\midrule
\multirow{5}{*}{\textit{ID-TAP Algorithm}}
& BT smoothing & $\alpha$ & 0.7 \\
& Acquisition floor & $\epsilon_g$ & $10^{-8}$ \\
& Softmax temperature & $\tau$ & 4.5 \\
& Sigmoid steepness & $\beta_{\text{sim}}$ & 6 \\
& Sigmoid midpoint & $\mu_{\text{sim}}$ & 0.8 \\
\midrule
\multirow{6}{*}{\textit{LLM Settings}}
& Model & -- & DeepSeek-V3.2 \\
& Temperature (generation) & -- & 0.6 \\
& Temperature (judge) & -- & 0.0 \\
& Max output tokens & -- & 1024 \\
& Frequency penalty & -- & 0.5 \\
& Presence penalty & -- & 0.5 \\
\bottomrule
\end{tabular}
\end{table*}

}

\FloatBarrier
\section{Persona-Construction and Density Analysis}
\label{app:persona_density_analysis}

\subsection{Construction and Exploration Diagnostic}

For each dataset, we compare iterative rephrasing (IR) with keyword-conditioned construction (KW) using a common candidate budget, response generator, embedding model, interaction budget, and evaluation protocol. IR follows the persona rephrasing procedure in Section~\ref{app:prompts}, including the reset interval $R=15$. KW extracts preference and style keywords from the same user history, assigns keyword combinations to candidate arms, and generates a persona conditioned on each combination. Both constructions use Qwen3-Embedding-4B representations before online optimization.

For method $m$, let $\mathcal{H}_{t-1}^{q,m}=\{p^{q,m}_{1,2},\ldots,p^{q,m}_{t-1,2}\}$ denote the challengers selected before round $t$. The Arm-2 exploration rate is
\begin{equation}
\operatorname{ExplorationRate}(m)
=\frac{1}{T-1}\sum_{t=2}^{T}
\mathbb{I}\!\left[p^{q,m}_{t,2}\notin\mathcal{H}_{t-1}^{q,m}\right].
\label{eq:arm2_exploration_rate}
\end{equation}
The first round is excluded because its challenger is necessarily unseen. A repeated challenger is counted as exploitation, and the same tie-handling implementation is used for all methods.

\begin{table}[H]
\centering
\caption{Mean pairwise cosine similarity under iterative rephrasing (IR) and keyword-conditioned construction (KW).}
\label{tab:ir_kw_density}
\small
\begin{tabular}{lrrr}
\toprule
Dataset & $\bar{\mu}$ (IR) & $\bar{\mu}$ (KW) & $\Delta\bar{\mu}$ \\
\midrule
LaMP-4     & 0.672 & 0.720 & $+0.048$ \\
LongLaMP-4 & 0.686 & 0.695 & $+0.009$ \\
PrefEval   & 0.716 & 0.830 & $+0.114$ \\
UltraChat  & 0.715 & 0.834 & $+0.119$ \\
\bottomrule
\end{tabular}
\end{table}

\begin{table}[H]
\centering
\caption{Arm-2 exploration rates under iterative rephrasing (IR).}
\label{tab:ir_exploration}
\small
\begin{tabular}{lrrr}
\toprule
Dataset & POHF & PDO Samp. & ID-TAP \\
\midrule
LaMP-4     & 24.1 & 25.7 & 73.5 \\
LongLaMP-4 & 25.3 & 28.1 & 72.2 \\
PrefEval   & 65.0 & 64.0 & 95.3 \\
UltraChat  & 67.3 & 66.8 & 94.8 \\
\bottomrule
\end{tabular}
\end{table}

\begin{table}[H]
\centering
\caption{Arm-2 exploration rates under keyword-conditioned construction (KW). Parentheses report the change from IR in percentage points.}
\label{tab:kw_exploration}
\small
\begin{tabular}{lrrr}
\toprule
Dataset & POHF & PDO Samp. & ID-TAP \\
\midrule
LaMP-4     & 16.6 ($-7.5$) & 17.6 ($-8.1$) & 74.8 ($+1.3$) \\
LongLaMP-4 & 16.5 ($-8.8$) & 20.3 ($-7.8$) & 71.9 ($-0.3$) \\
PrefEval   & 61.0 ($-4.0$) & 62.5 ($-1.5$) & 94.4 ($-0.9$) \\
UltraChat  & 64.8 ($-2.5$) & 60.5 ($-6.3$) & 93.3 ($-1.5$) \\
\bottomrule
\end{tabular}
\end{table}

Across all four datasets, KW increases mean persona similarity and is accompanied by lower exploration for POHF and PDO Sampling. ID-TAP's exploration rate changes by at most 1.5 percentage points between the two constructions. Together, these results show an association between denser persona spaces and reduced conventional exploration, alongside comparatively stable exploration by ID-TAP.

\subsection{Exploration--Performance Association}

\begin{table}[H]
\centering
\caption{Association between conventional exploration under KW and ID-TAP's ROUGE-L margin over the strongest baseline.}
\label{tab:exploration_performance_association}
\small
\resizebox{\columnwidth}{!}{%
\begin{tabular}{lrr}
\toprule
Dataset group & Mean POHF exploration & Mean ID-TAP margin \\
\midrule
LaMP-4 and LongLaMP-4 & 16.55\% & $+0.0160$ \\
PrefEval and UltraChat & 62.90\% & $+0.0065$ \\
\bottomrule
\end{tabular}
}
\end{table}

The datasets with lower POHF exploration under KW also have the larger reported ID-TAP margin. Together with the within-dataset IR--KW comparison, this descriptive pattern aligns lower conventional exploration with a larger performance margin for ID-TAP.

\subsection{Sensitivity to Persona-Space Size and Construction}

In the following tables, $n_{\mathrm{user}}=40$ denotes sampled users per dataset. These experiments use $K=100$ candidates, whereas the main experiments use $K=400$.

\begin{table}[H]
\centering
\caption{ROUGE-L with iterative-rephrasing personas ($K=100$, $n_{\mathrm{user}}=40$ sampled users per dataset).}
\label{tab:k100_ir_results}
\small
\resizebox{\columnwidth}{!}{%
\begin{tabular}{lrrrr}
\toprule
Dataset & ID-TAP & POHF & PDO Samp. & PersonaAgent \\
\midrule
UltraChat   & \textbf{0.2430} & 0.2291 & 0.2213 & 0.1057 \\
PrefEval    & \textbf{0.2587} & 0.2498 & 0.2430 & 0.1651 \\
LaMP-4      & \textbf{0.2262} & 0.1900 & 0.2001 & 0.1529 \\
LongLaMP-4  & \textbf{0.1392} & 0.1310 & 0.1320 & 0.1230 \\
\bottomrule
\end{tabular}
}
\end{table}

\begin{table}[H]
\centering
\caption{ROUGE-L with keyword-conditioned personas ($K=100$, $n_{\mathrm{user}}=40$ sampled users per dataset).}
\label{tab:k100_kw_results}
\small
\resizebox{\columnwidth}{!}{%
\begin{tabular}{lrrrr}
\toprule
Dataset & ID-TAP & POHF & PDO Samp. & PersonaAgent \\
\midrule
UltraChat   & \textbf{0.2565} & 0.2441 & 0.2496 & 0.1090 \\
PrefEval    & \textbf{0.2614} & 0.2556 & 0.2527 & 0.1647 \\
LaMP-4      & \textbf{0.2495} & 0.2201 & 0.2172 & 0.1554 \\
LongLaMP-4  & \textbf{0.1388} & 0.1356 & 0.1350 & 0.1253 \\
\bottomrule
\end{tabular}
}
\end{table}

At $K=100$, ID-TAP achieves the highest mean ROUGE-L under both candidate-space constructions, extending the main $K=400$ result to a smaller persona pool and an alternative construction procedure.

\section{Robustness to Preference and Evaluation Models}
\label{app:robustness}

We evaluate three settings: GPT-4o for both preference generation and final evaluation, GPT-4o for preference generation with DeepSeek-V3.2 for final evaluation, and BERTScore as an evaluation metric independent of the optimizer's preference signal. Each setting follows the main protocol for candidate pools, interaction budgets, response generation, prompts, and aggregation, while varying the specified preference or evaluation component.

\begin{table}[H]
\centering
\caption{Win rate with GPT-4o used for both preference generation and evaluation.}
\label{tab:gpt4o_matched}
\small
\resizebox{\columnwidth}{!}{%
\begin{tabular}{lrrrr}
\toprule
Dataset & PersonaAgent & POHF & PDO Samp. & ID-TAP \\
\midrule
LaMP-4    & 73.2 & 80.2 & 80.4 & \textbf{81.9} \\
UltraChat & 81.7 & 86.7 & 85.8 & \textbf{89.1} \\
\bottomrule
\end{tabular}
}
\end{table}

\begin{table}[H]
\centering
\caption{Win rate with GPT-4o preferences and DeepSeek-V3.2 evaluation.}
\label{tab:gpt4o_deepseek_mismatch}
\small
\resizebox{\columnwidth}{!}{%
\begin{tabular}{lrrrr}
\toprule
Dataset & PersonaAgent & POHF & PDO Samp. & ID-TAP \\
\midrule
LaMP-4    & 68.1 & 75.0 & 77.2 & \textbf{78.2} \\
UltraChat & 64.2 & 85.0 & 82.5 & \textbf{89.2} \\
\bottomrule
\end{tabular}
}
\end{table}

\begin{table}[H]
\centering
\caption{BERTScore under a metric independent of the preference-optimization signal.}
\label{tab:bertscore_robustness}
\small
\resizebox{\columnwidth}{!}{%
\begin{tabular}{lrrrr}
\toprule
Dataset & PersonaAgent & POHF & PDO Samp. & ID-TAP \\
\midrule
UltraChat  & 0.838 & 0.890 & 0.889 & \textbf{0.892} \\
PrefEval   & 0.847 & 0.885 & 0.887 & \textbf{0.888} \\
LaMP-4     & 0.862 & 0.870 & 0.872 & \textbf{0.876} \\
LongLaMP-4 & 0.812 & 0.814 & 0.812 & \textbf{0.818} \\
\bottomrule
\end{tabular}
}
\end{table}

ID-TAP achieves the highest reported mean in every configuration. The matched GPT-4o setting changes the automated preference source and final evaluator together, the GPT-4o/DeepSeek-V3.2 setting separates those roles, and BERTScore evaluates responses independently of the preference-optimization signal. The resulting ordering is consistent across the three evaluation configurations.

\section{Runtime and Timing Protocol}
\label{app:runtime}

Timing measurements were collected on a workstation equipped with a single NVIDIA GeForce RTX 4090 GPU using the model-serving backend, batch behavior, and concurrency configuration of the main experiments. Here, $K$ denotes the number of candidate personas in each user's persona space, and $n_{\mathrm{user}}$ denotes the number of sampled users per dataset. The measurements use $K=100$ and $n_{\mathrm{user}}=40$. Each user's timing is averaged over three independent runs, and the tables report the mean, standard deviation, and SEM across these per-user averages. Wall-clock time covers a complete interaction round, including response generation and evaluation; update time isolates persona selection, BT-MLE estimation, neural-model updating, and entropy-proxy computation.

\begin{table}[H]
\centering
\caption{ID-TAP wall-clock time with $K=100$ and $n_{\mathrm{user}}=40$ sampled users per dataset. Statistics are computed across per-user averages of three independent runs and reported in seconds per interaction round.}
\label{tab:runtime_wall}
\small
\begin{tabular}{lrrr}
\toprule
Dataset & Mean & Std. & SEM \\
\midrule
UltraChat  & 1.1142 & 0.4211 & 0.0666 \\
PrefEval   & 0.9513 & 0.1745 & 0.0276 \\
LaMP-4     & 2.5674 & 0.9069 & 0.1434 \\
LongLaMP-4 & 3.1499 & 0.9750 & 0.1542 \\
\bottomrule
\end{tabular}
\end{table}

\begin{table}[H]
\centering
\caption{ID-TAP algorithmic update time with $K=100$ and $n_{\mathrm{user}}=40$ sampled users per dataset. Statistics are computed across per-user averages of three independent runs and reported in seconds per interaction round.}
\label{tab:runtime_update}
\small
\begin{tabular}{lrrr}
\toprule
Dataset & Mean & Std. & SEM \\
\midrule
UltraChat  & 0.0964 & 0.0189 & 0.0030 \\
PrefEval   & 0.0651 & 0.0157 & 0.0025 \\
LaMP-4     & 0.5709 & 0.1459 & 0.0231 \\
LongLaMP-4 & 0.3172 & 0.0839 & 0.0133 \\
\bottomrule
\end{tabular}
\end{table}

Algorithmic updates require 0.065--0.571 seconds per round, compared with 0.95--3.15 seconds end to end. Based on the reported means, the algorithmic component accounts for approximately 7--22\% of total latency, with model serving dominating the measured wall-clock time.

\section{Persona Examples}
\label{app:persona_examples}

\tcbset{
  personabox/.style={
    enhanced,
    boxrule=0.5pt,
    colframe=black,
    colback=white,
    colbacktitle=white,
    coltitle=black,
    fonttitle=\bfseries\large,
    halign title=center,
    titlerule=0.4pt,
    top=2mm,
    bottom=2mm,
    left=2mm,
    right=2mm,
    toptitle=1.5mm,
    bottomtitle=1.5mm
  }
}

\subsection{LaMP-4}

\begin{tcolorbox}[personabox, title=Persona A]
\footnotesize
This user approaches political coverage as a strategic observer, with a particular focus
on the mechanics of the 2016 Republican primary. Their analytical lens is drawn to the
concrete maneuvers and positional conflicts between figures like Donald Trump, Ted Cruz,
and Jeb Bush. They consistently select content detailing the ``neck-and-neck battle'' in
Iowa or the logistical specifics of a vice-presidential search, prioritizing articles that
dissect how candidates compete. Their neutral tone and preference for headlines framing
events as consequential clashes---a ``fiery debate'' or a ``growing feud''---reveal a
perspective that treats politics as a complex game. The core interest remains in tracking
concrete developments, such as poll results and campaign strategies that signal who might
``knock out'' rivals.
\end{tcolorbox}

\begin{tcolorbox}[personabox, title=Persona B]
\footnotesize
This user approaches political campaigns as intricate systems to be reverse-engineered,
with a focus on how discrete operational choices cascade into winning or losing outcomes.
Their detailed study of the 2016 Republican primary, particularly the delegate calculations
that defined the Trump--Cruz dynamic in Iowa, serves as a foundational case. They treat
strategic events---like Marco Rubio's decision to decline a vice-presidential spot---not as
political drama but as pivotal system inputs that reroute a campaign's entire pathway.
Polling shifts and tactical proposals are analyzed as interconnected variables within a
dynamic model.
\end{tcolorbox}

\begin{tcolorbox}[personabox, title=Persona C]
\footnotesize
The user is a politically observant individual with a concentrated focus on the 2016
Republican presidential primary's competitive landscape. Their interest centers on key
figures like Donald Trump, Ted Cruz, Jeb Bush, and Marco Rubio, with attention to pivotal
moments such as Trump's vice-presidential search involving Mike Pence and the
``neck-and-neck battle'' in Iowa polls. They consistently favor content analyzing electoral
mechanics and candidate positioning---tracking how a Trump sweep could ``knock out'' rivals
in Florida and Ohio. Their chosen material frames events through a lens of conflict and
consequence, highlighting a ``growing feud,'' a ``fiery debate,'' or campaigns
``scrambling to make up ground.''
\end{tcolorbox}

\noindent\begin{minipage}{\linewidth}

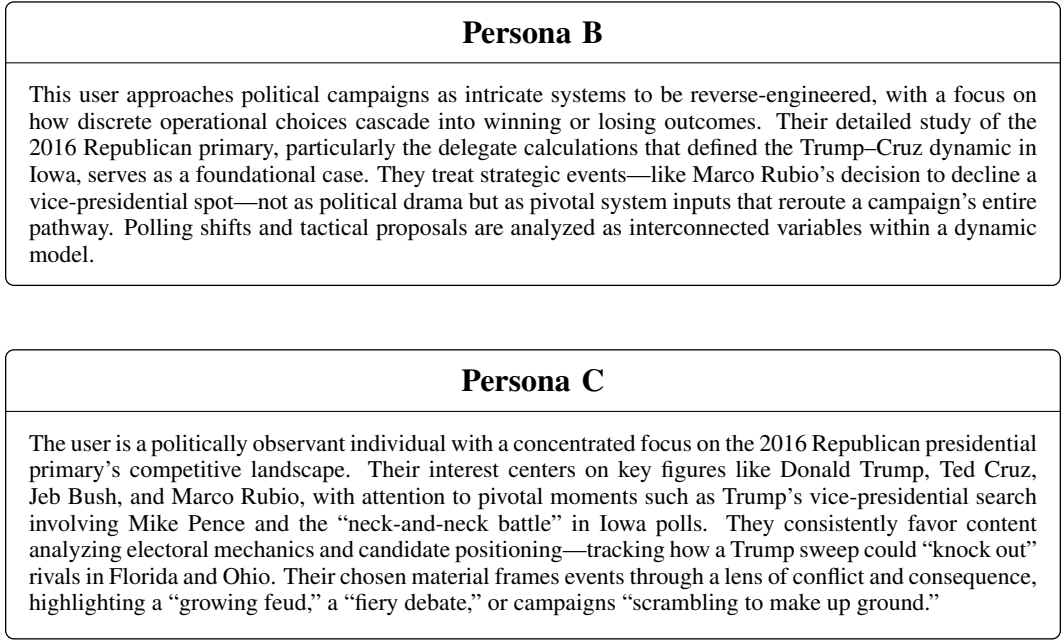
\captionof{figure}{Three persona descriptions sampled from the same user domain in
\textbf{LaMP-4} (news headline generation), illustrating the diversity of summaries
maintained in the persona bank.}
\end{minipage}

\subsection{LongLaMP-4}

\begin{tcolorbox}[personabox, title=Persona A]
\footnotesize
This user is a systematic thinker who values logical consistency, clear hierarchies, and
functional design across diverse domains. They critique \textit{Minecraft}'s development
for lacking direction, advocating for a structured ``wood--stone--iron--diamond'' material
system over random additions like the ``bizarre endgame'' and poorly implemented potions.
Their analytical approach extends to physics, explaining complex concepts like 4-vectors
by simplifying them into core principles. They exhibit strong, evidence-backed opinions,
defending financial markets as providing liquidity and downward price pressure, while
dismissing overly broad definitions of art. Their communication is assertive and detailed,
often using personal comparisons---like favoring a \$700 HP laptop over a \$1500
MacBook---to ground arguments in pragmatic cost-benefit analysis. This pattern reveals a
persona deeply invested in deconstructing systems, prioritizing structured understanding
and practical utility over abstract or sentimental reasoning.
\end{tcolorbox}

\begin{tcolorbox}[personabox, title=Persona B]
\footnotesize
This thinker approaches systems with a forensic eye for functional hierarchy and practical
coherence. Their satisfaction comes from mapping how components interlock to produce
reliable outcomes, whether in game design's reward loops or technology's
price-to-performance ratios. They defend \textit{Minecraft}'s clean
``wood--stone--iron--diamond'' progression as logically sound architecture, critiquing
later potion systems as messy deviations. Their analysis of consumer tech, such as
comparing a \$700 HP laptop to a \$1500 MacBook, seeks tangible benchmarks over brand
prestige. Communication is direct and dismantling: they prefer explaining complex
phenomena---like the constancy of light via 4-vectors or financial instruments as liquidity
tools---through grounded mechanics. For them, rigor is the removal of sentimental
abstraction to reveal the operative skeleton beneath.
\end{tcolorbox}

\begin{tcolorbox}[personabox, title=Persona C]
\footnotesize
This user is fundamentally a model-builder, driven to synthesize disparate domains into
coherent, actionable frameworks. Their analytical skill is applied to practical puzzles,
from evaluating laptop specifications to dissecting \textit{Minecraft} enchantment systems,
treating each as a case study in translating convoluted rules into functional logic. While
conversant with advanced concepts like 4-vectors, their true talent lies in
integration---weaving threads from art criticism and financial theory to construct robust
mental models. Their confidence is pragmatic, rooted in the belief that more elegant,
integrated systems are not merely ideal but essential for progress. They operate as a
pragmatic architect of knowledge, specifically focused on the intersection where
technological systems meet economic principles.
\end{tcolorbox}

\noindent\begin{minipage}{\linewidth}
\captionof{figure}{Three persona descriptions sampled from the same user domain in
\textbf{LongLaMP-4} (topic-focused long-form writing). The user domain spans game design
critique, consumer technology analysis, physics exposition, and economic commentary.}
\end{minipage}

\clearpage
\section{More Experimental Results}
\label{app:additional_main_results}

Figure~\ref{fig:query_all} reports the individual-dataset results underlying the aggregate view in the main paper. Figure~\ref{fig:itr_all} shows how performance evolves within the fixed comparison budget and provides dataset-specific detail beyond the aggregate curve.

\subsection{Detailed Performance Comparison Across Datasets}
\label{app:performance_comparison}

\begin{figure}[H]
    \centering
    \begin{subfigure}[t]{0.30\textwidth}
        \centering
        \includegraphics[width=\linewidth]{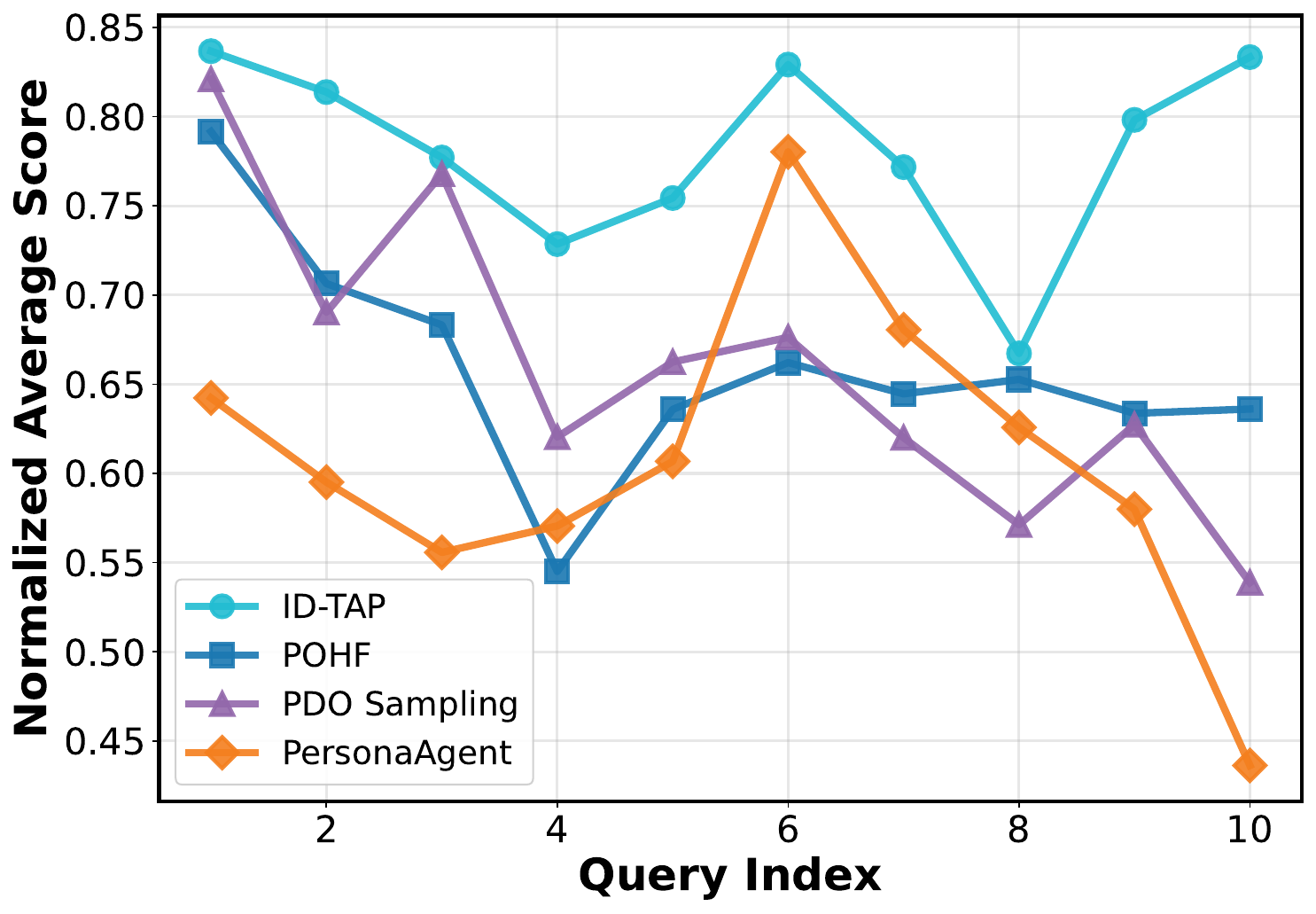}
        \caption{LaMP-4: News Headline}
        \label{fig:query_lamp4}
    \end{subfigure}
    \hspace{0.025\textwidth}
    \begin{subfigure}[t]{0.30\textwidth}
        \centering
        \includegraphics[width=\linewidth]{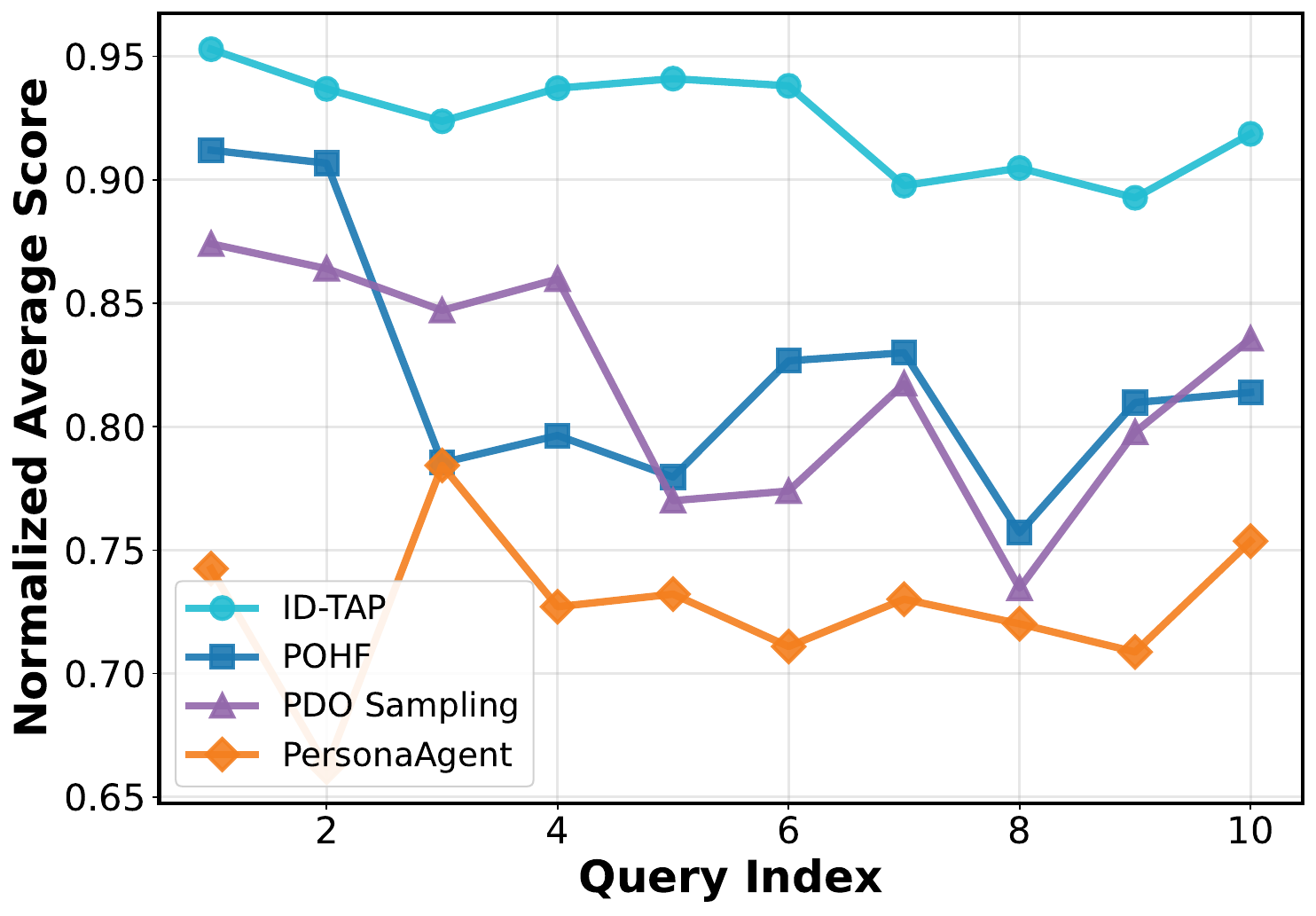}
        \caption{LaMP-5: Scholarly Title}
        \label{fig:query_lamp5}
    \end{subfigure}
    \par\smallskip
    \begin{subfigure}[t]{0.30\textwidth}
        \centering
        \includegraphics[width=\linewidth]{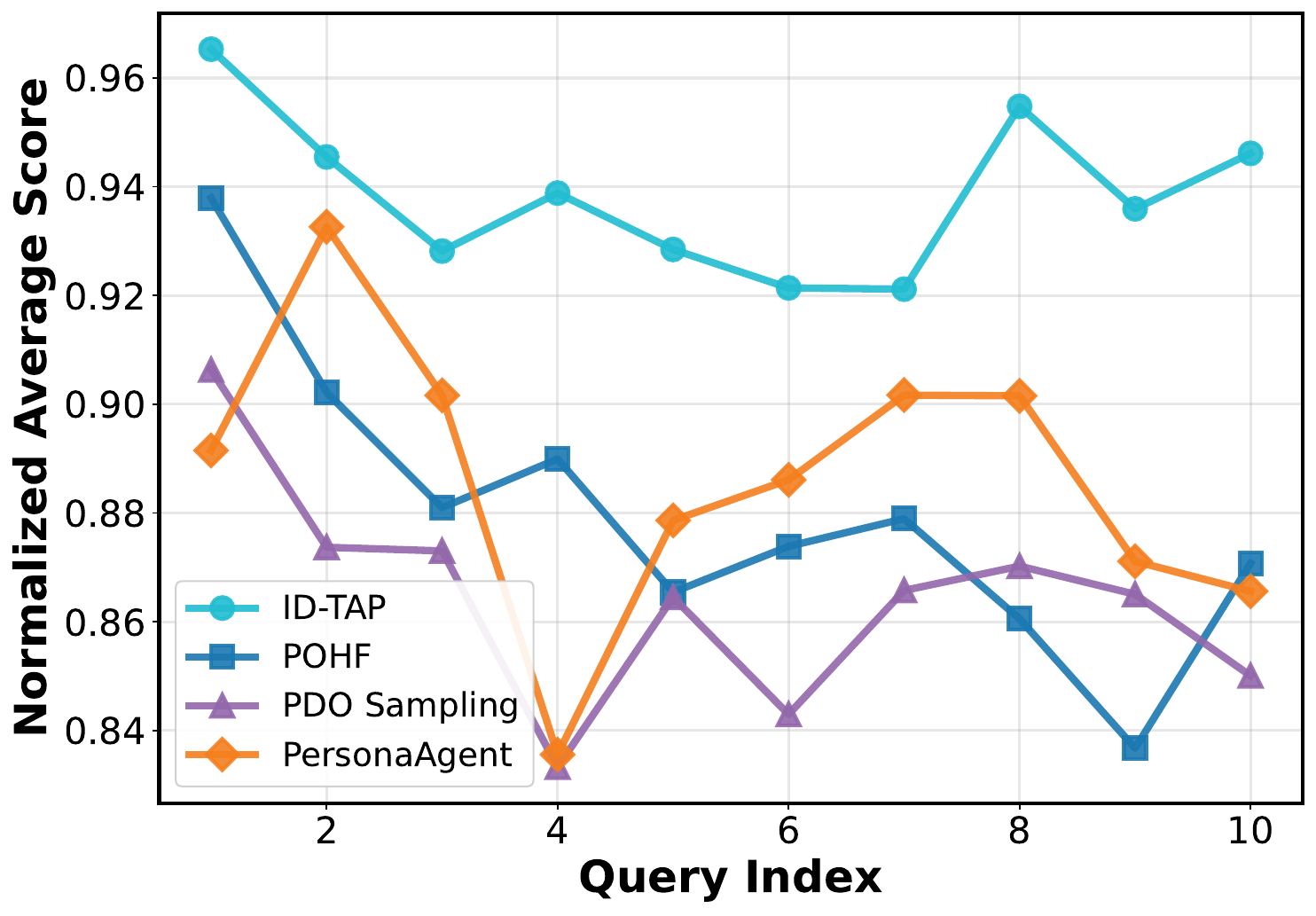}
        \caption{LongLaMP-2: Abstract}
        \label{fig:query_lamp8}
    \end{subfigure}
    \hfill
    \begin{subfigure}[t]{0.30\textwidth}
        \centering
        \includegraphics[width=\linewidth]{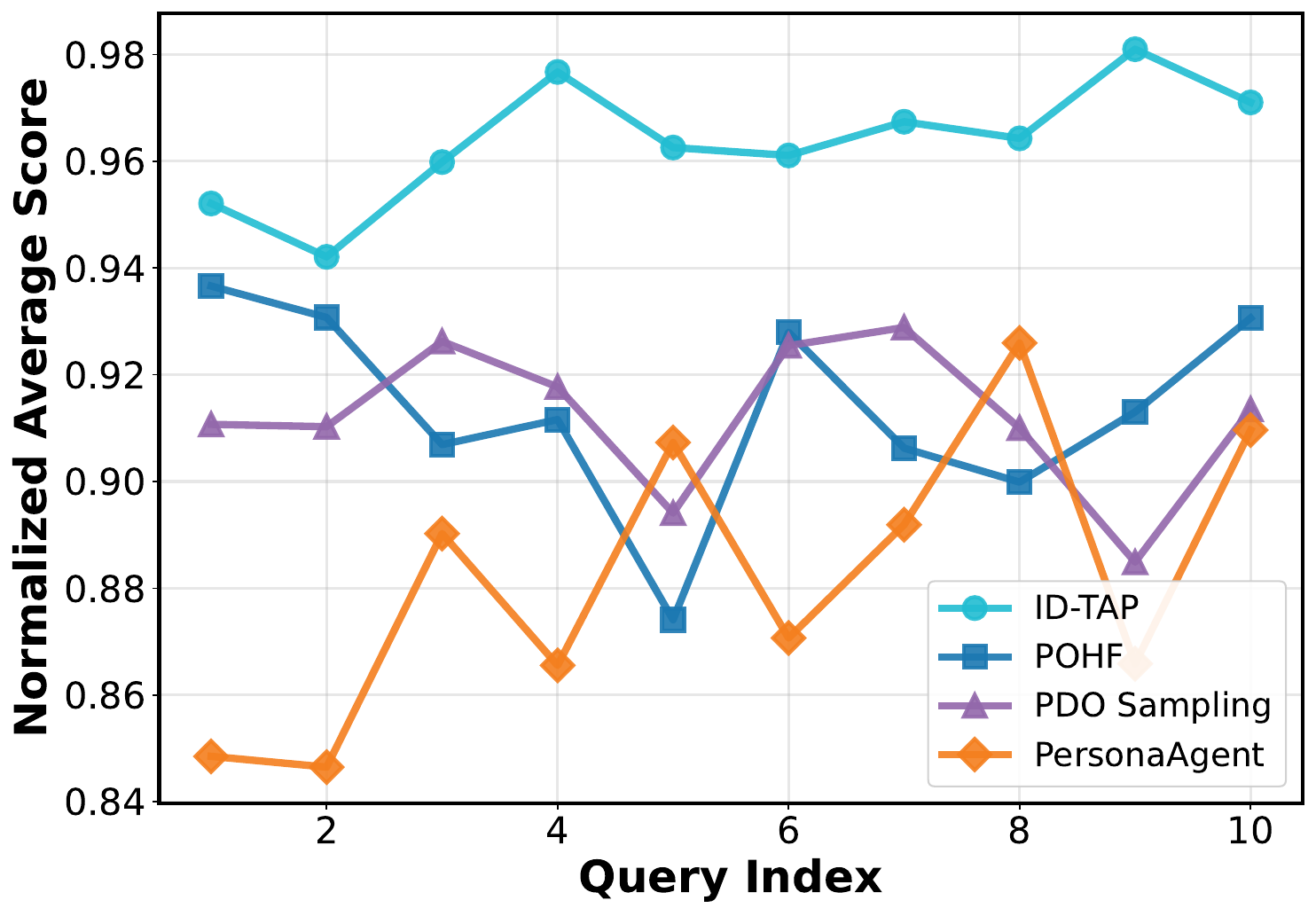}
        \caption{LongLaMP-3: Review}
        \label{fig:query_lamp9}
    \end{subfigure}
    \hfill
    \begin{subfigure}[t]{0.30\textwidth}
        \centering
        \includegraphics[width=\linewidth]{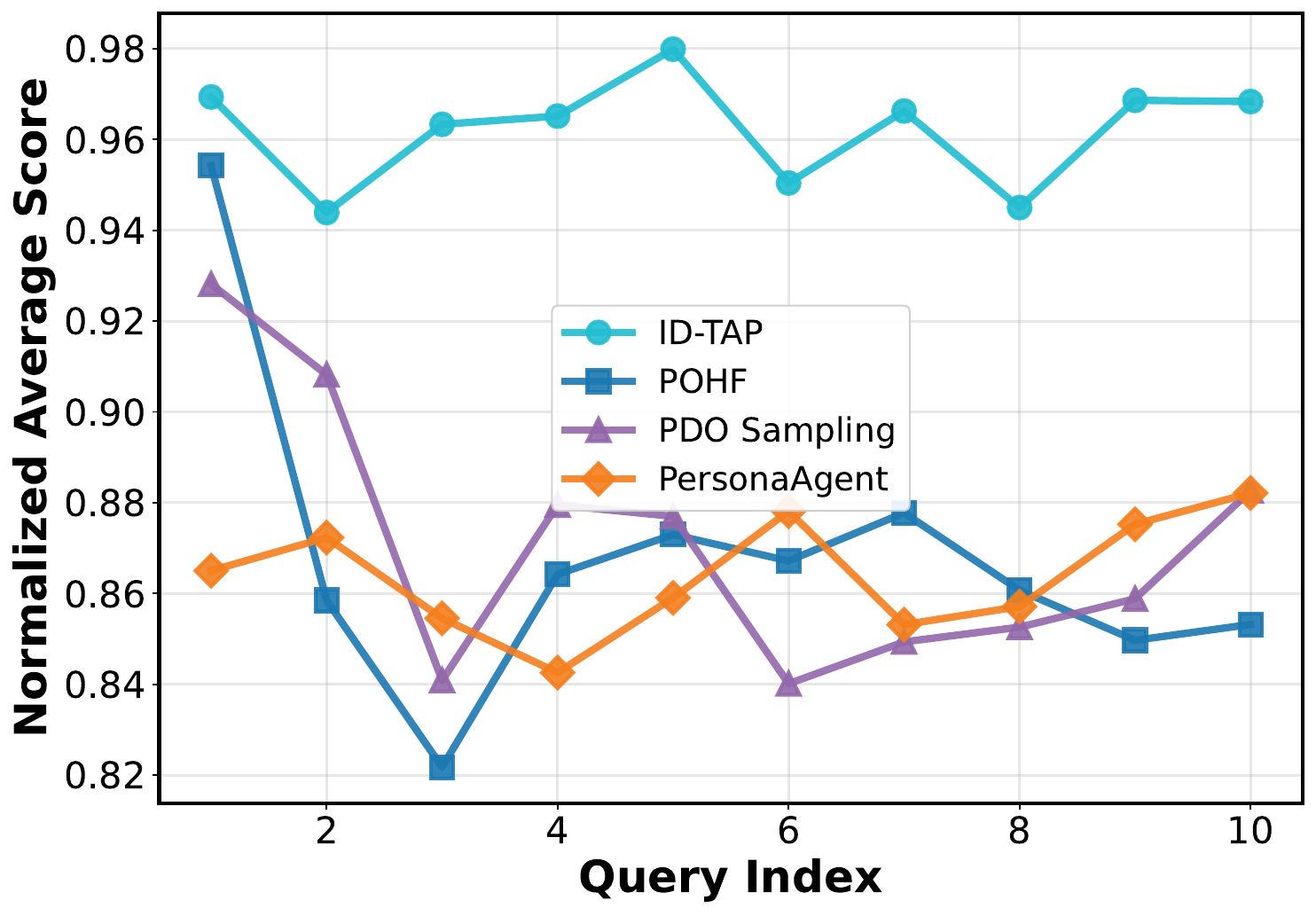}
        \caption{LongLaMP-4: Topic}
        \label{fig:query_lamp10}
    \end{subfigure}
    \par\smallskip
    \begin{subfigure}[t]{0.30\textwidth}
        \centering
        \includegraphics[width=\linewidth]{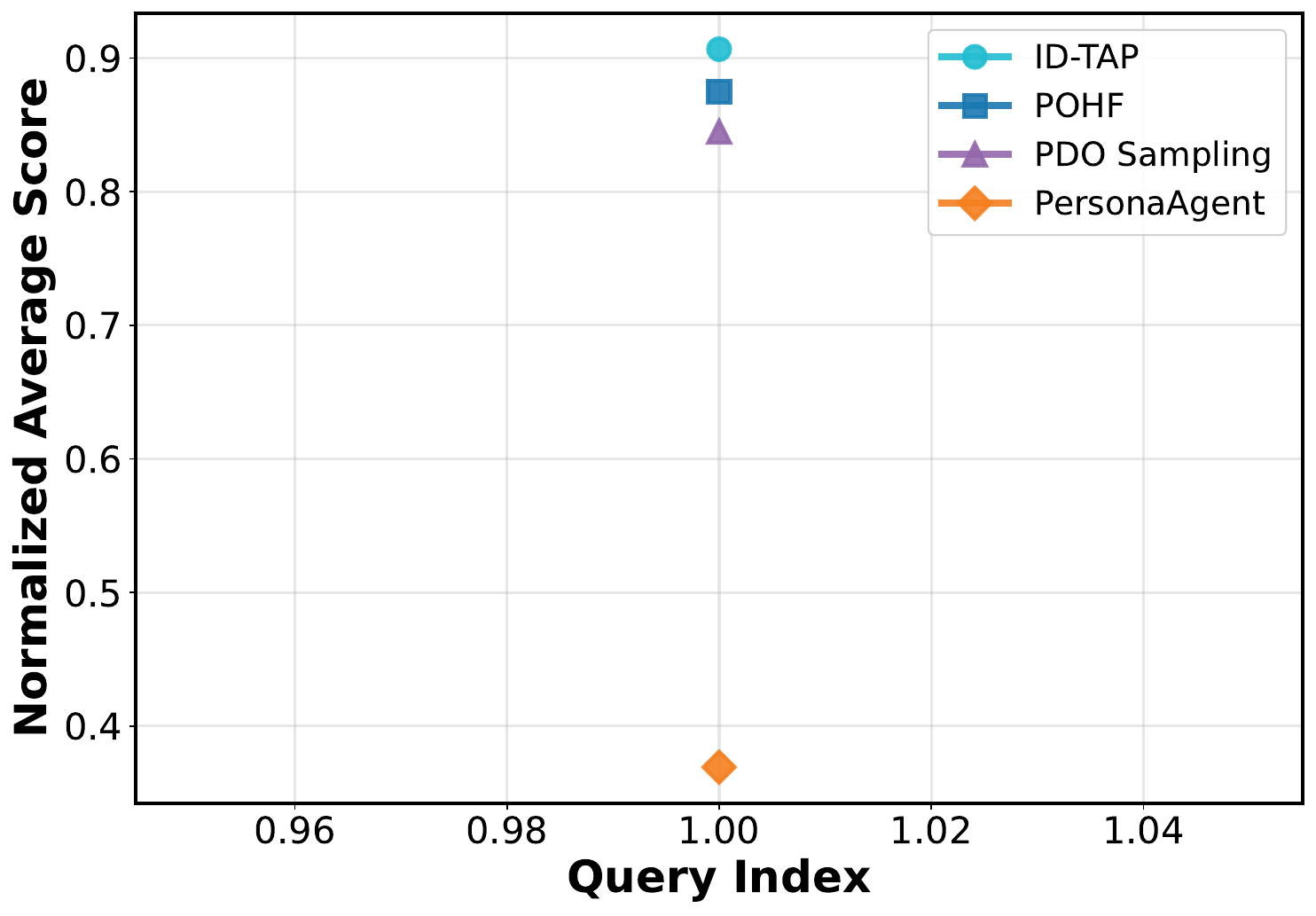}
        \caption{UltraChat}
        \label{fig:query_ultrachat}
    \end{subfigure}
    \hspace{0.025\textwidth}
    \begin{subfigure}[t]{0.30\textwidth}
        \centering
        \includegraphics[width=\linewidth]{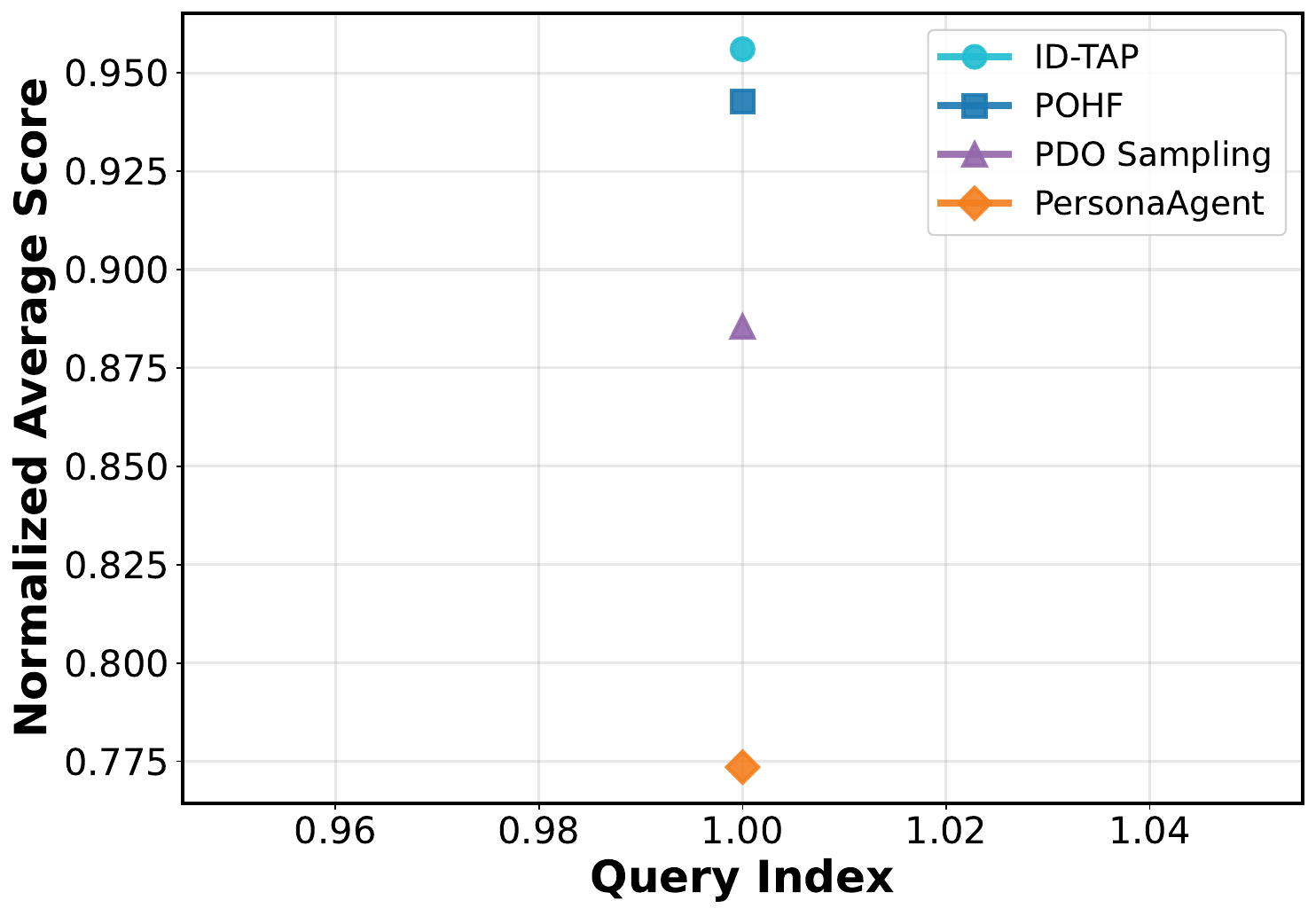}
        \caption{PrefEval}
        \label{fig:query_prefeval}
    \end{subfigure}
    \caption{Query-level performance across all evaluated benchmarks.}
    \label{fig:query_all}
\end{figure}

Figure~\ref{fig:query_all} compares ID-TAP across three benchmark categories: (1) \textbf{LaMP}, covering personalized headline and scholarly-title generation; (2) \textbf{LongLaMP}, covering personalized abstract, review, and topic generation; and (3) \textbf{multi-turn dialogue}, covering UltraChat and PrefEval.

\subsection{Detailed Iteration-Level Trajectories}
\label{app:iteration_trajectories}

\begin{figure}[H]
    \centering
    \begin{subfigure}[t]{0.30\textwidth}
        \centering
        \includegraphics[width=\linewidth]{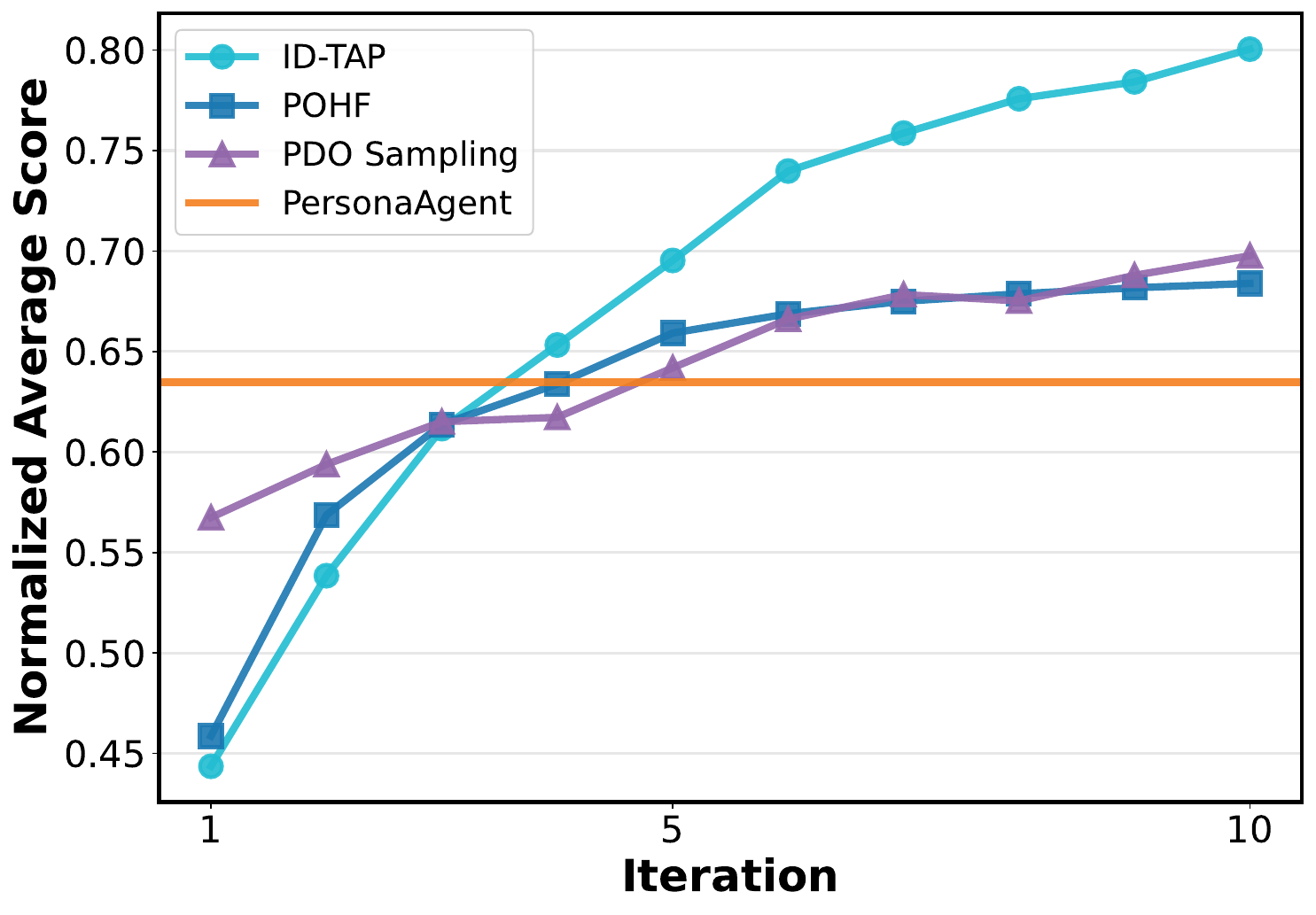}
        \caption{LaMP-4: News Headline}
        \label{fig:itr_lamp4}
    \end{subfigure}
    \hspace{0.025\textwidth}
    \begin{subfigure}[t]{0.30\textwidth}
        \centering
        \includegraphics[width=\linewidth]{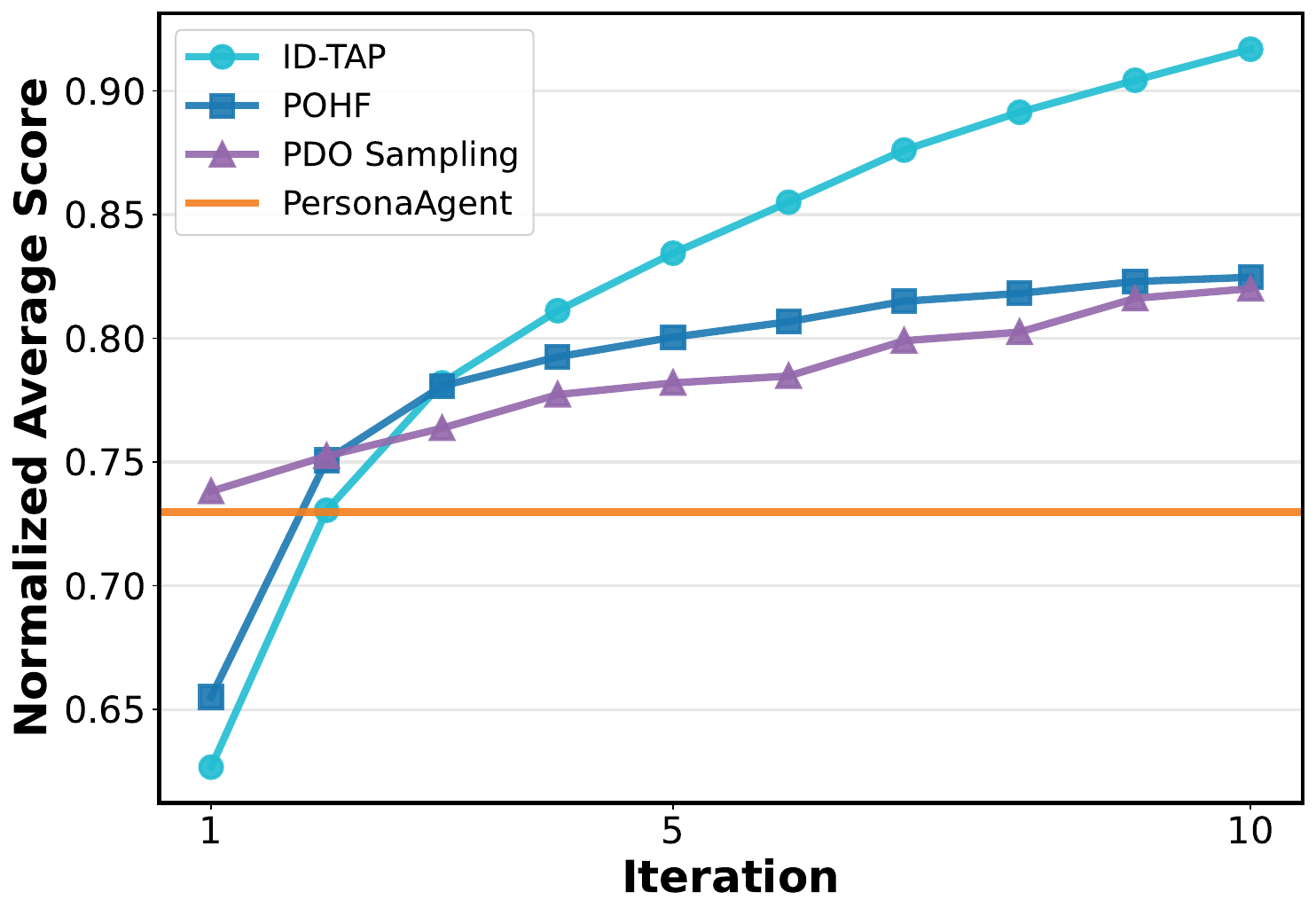}
        \caption{LaMP-5: Scholarly Title}
        \label{fig:itr_lamp5}
    \end{subfigure}
    \par\smallskip
    \begin{subfigure}[t]{0.30\textwidth}
        \centering
        \includegraphics[width=\linewidth]{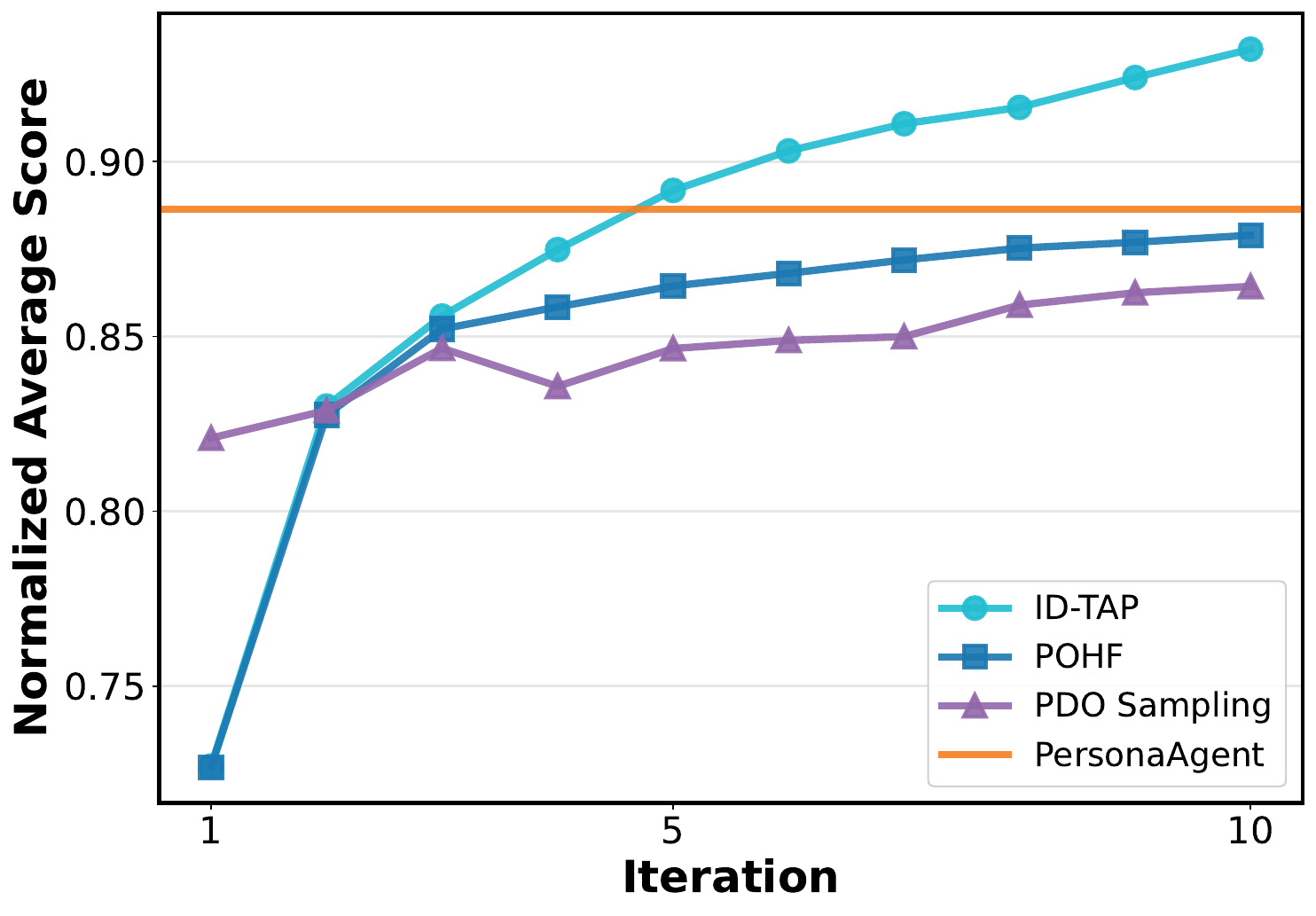}
        \caption{LongLaMP-2: Abstract}
        \label{fig:itr_lamp8}
    \end{subfigure}
    \hfill
    \begin{subfigure}[t]{0.30\textwidth}
        \centering
        \includegraphics[width=\linewidth]{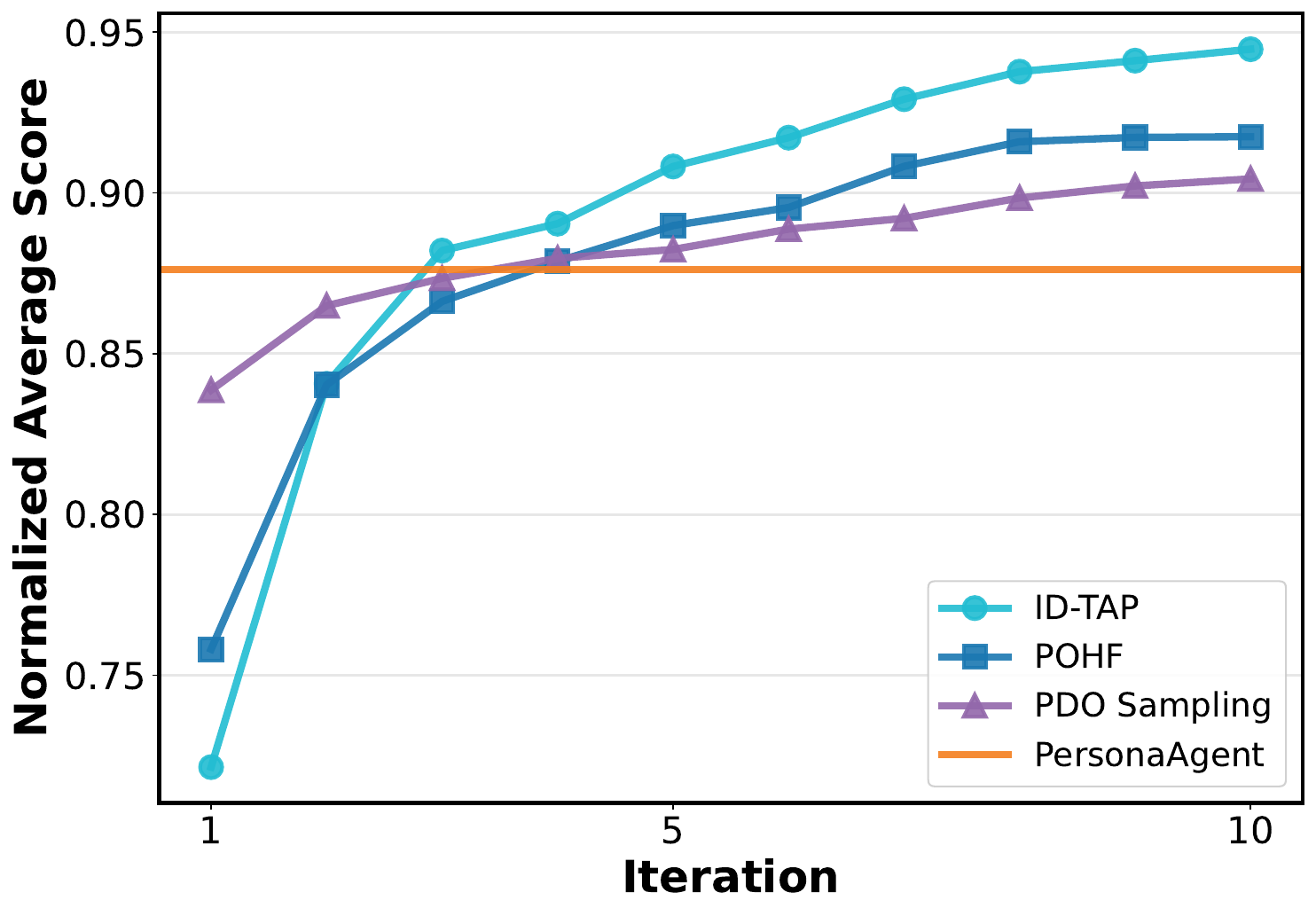}
        \caption{LongLaMP-3: Review}
        \label{fig:itr_lamp9}
    \end{subfigure}
    \hfill
    \begin{subfigure}[t]{0.30\textwidth}
        \centering
        \includegraphics[width=\linewidth]{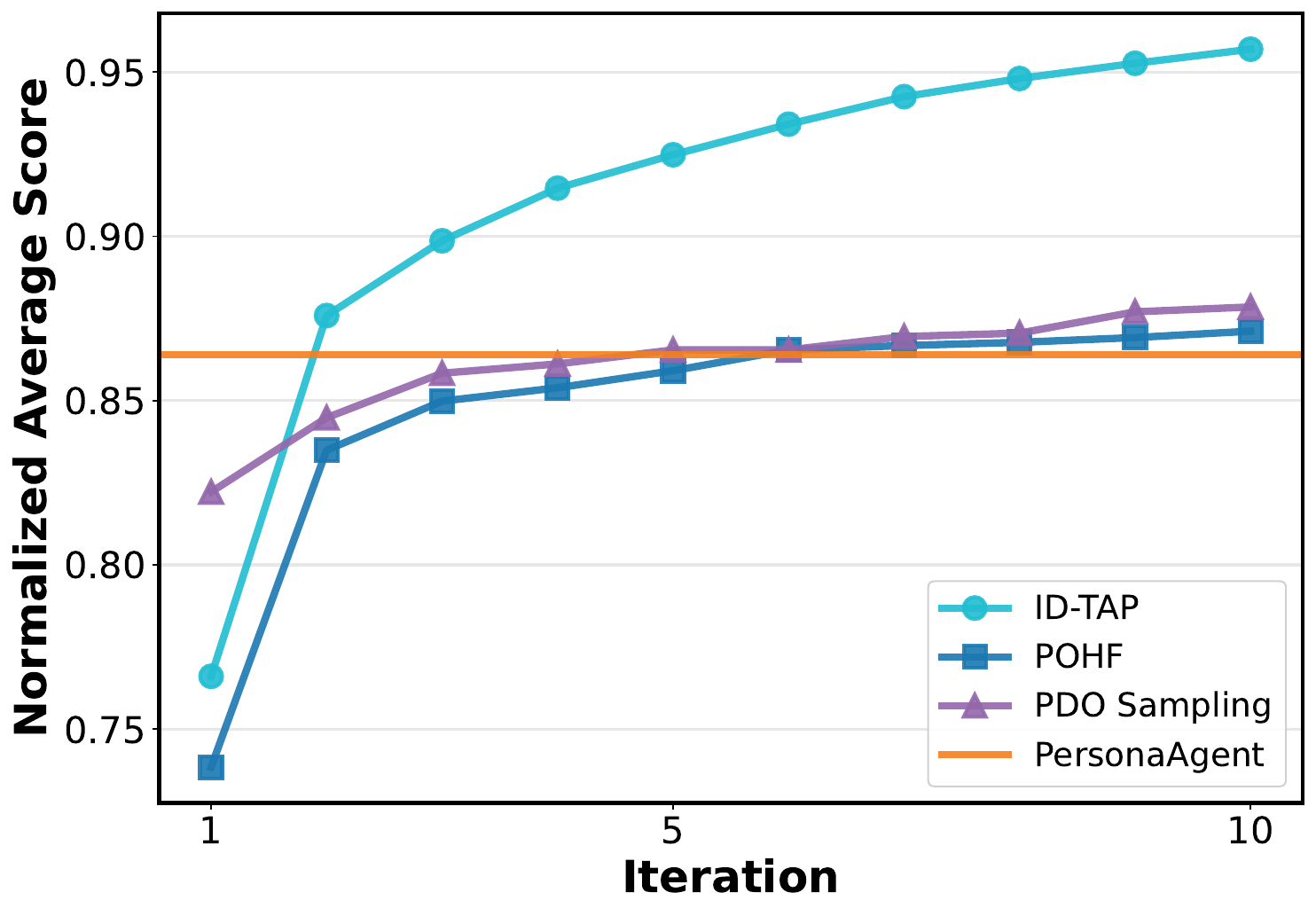}
        \caption{LongLaMP-4: Topic}
        \label{fig:itr_lamp10}
    \end{subfigure}
    \par\smallskip
    \begin{subfigure}[t]{0.30\textwidth}
        \centering
        \includegraphics[width=\linewidth]{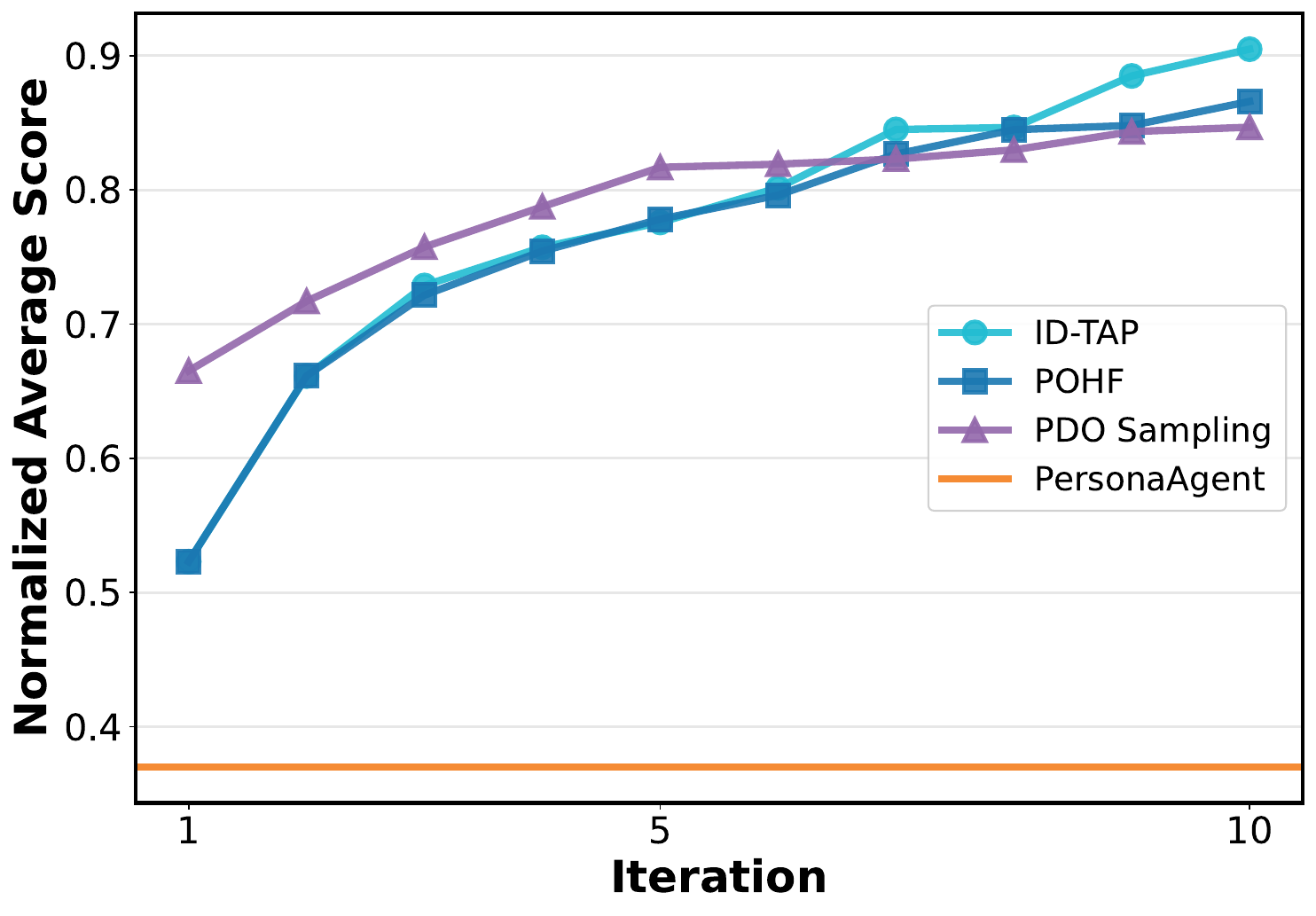}
        \caption{UltraChat}
        \label{fig:itr_ultrachat}
    \end{subfigure}
    \hspace{0.025\textwidth}
    \begin{subfigure}[t]{0.30\textwidth}
        \centering
        \includegraphics[width=\linewidth]{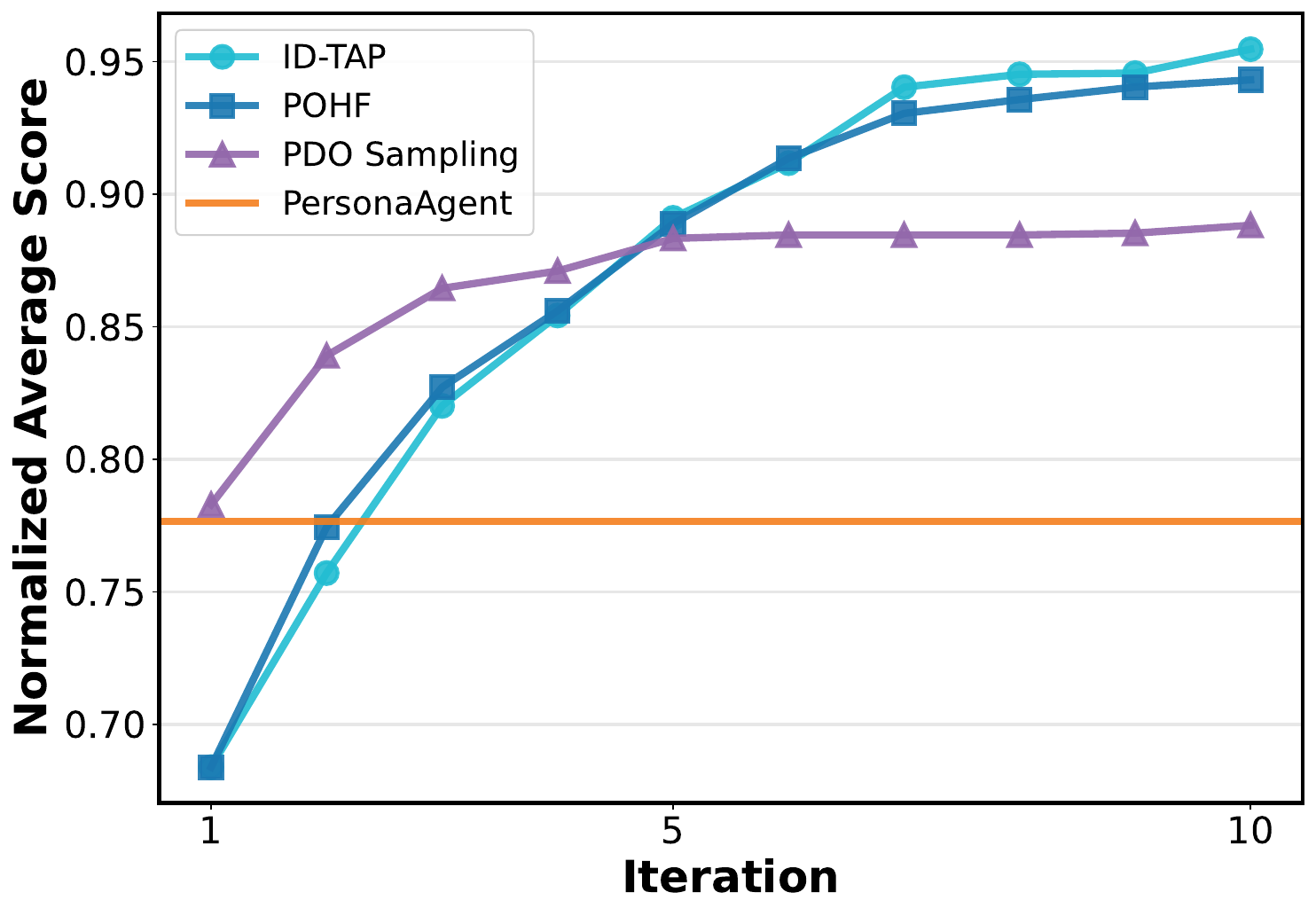}
        \caption{PrefEval}
        \label{fig:itr_prefeval}
    \end{subfigure}
    \caption{Iteration-level trajectories across all evaluated benchmarks.}
    \label{fig:itr_all}
\end{figure}

Figure~\ref{fig:itr_all} presents the iteration-level trajectory for each benchmark under automated proxy feedback.

\end{document}